\documentclass[lettersize,journal]{IEEEtran}
\usepackage{amsmath,amsfonts}
\usepackage{algorithmic}
\usepackage{algorithm}
\usepackage{array}
\usepackage{textcomp}
\usepackage{stfloats}
\usepackage{url}
\usepackage{verbatim}
\usepackage{graphicx}
\usepackage{cite}

\usepackage{enumitem}
\usepackage{subcaption}
\usepackage{multirow}
\usepackage{booktabs}
\usepackage{bbding}
\usepackage{fontawesome}
\usepackage{lineno, hyperref}

\begin{document}

\title{Scale-wise Bidirectional Alignment Network for Referring Remote Sensing Image Segmentation}

\author{Kun Li, George Vosselman, Michael Ying Yang
        % <-this % stops a space
\thanks{Kun Li and George Vosselman are with the Faculty of ITC, University of Twente, the Netherlands (k.li@utwente.nl; george.vosselman@utwente.nl).}% <-this % stops a space
\thanks{Michael Ying Yang is with the Visual Computing Group, Department of Computer Science, University of Bath, UK (myy35@bath.ac.uk).}
}

\markboth{Journal of \LaTeX\ Class Files, January~2025}%
{Shell \MakeLowercase{\textit{et al.}}: A Sample Article Using IEEEtran.cls for IEEE Journals}


\maketitle

\begin{abstract}
The goal of referring remote sensing image segmentation (RRSIS) is to extract specific pixel-level regions within an aerial image via a natural language expression.
% Recent methods have achieved impressive progress for RRSIS, including different Transformer-based fusion designs.
Recent advancements, particularly Transformer-based fusion designs, have demonstrated remarkable progress in this domain.
% However, existing methods only refine visual features through language-aware guidance in the cross-modal fusion stage, neglecting the opposite vision-to-language flow and resulting in irrelevant representations.
However, existing methods primarily focus on refining visual features using language-aware guidance during the cross-modal fusion stage, neglecting the complementary vision-to-language flow.
This limitation often leads to irrelevant or suboptimal representations.
% In addition, the complex spatial scales of various ground objects in aerial images challenge prior models' visual perception capability conditioned on the linguistic input.
In addition, the diverse spatial scales of ground objects in aerial images pose significant challenges to the visual perception capabilities of existing models when conditioned on textual inputs.
In this paper, we propose an innovative framework called Scale-wise Bidirectional Alignment Network (SBANet) to address these challenges for RRSIS.
Specifically, we design a Bidirectional Alignment Module (BAM) with learnable query tokens to selectively and effectively represent visual and linguistic features, emphasizing regions associated with key tokens.
% focusing on different regions related to crucial tokens.
BAM is further enhanced with a dynamic feature selection block, designed to provide both macro- and micro-level visual features, preserving global context and local details to facilitate more effective cross-modal interaction.
% that preserve global context and local details for better cross-modal interaction.
Furthermore, SBANet incorporates a text-conditioned channel and spatial aggregator to bridge the gap between the encoder and decoder, enhancing cross-scale information exchange in complex aerial scenarios.
% and improve the cross-scale information exchange in aerial scenarios.
Extensive experiments demonstrate that our proposed method achieves superior performance in comparison to previous state-of-the-art methods on the RRSIS-D and RefSegRS datasets, both quantitatively and qualitatively.
The code will be released after publication.
\end{abstract}


\section{Introduction}
\label{sec: intro}
\IEEEPARstart{R}{eferring} image segmentation (RIS) aims to segment a target object within a given image based on a natural language expression.
% Different from traditional single-modality image segmentation that assigns pre-defined labels to all pixels, RIS requires cross-modal understanding and locating referent at pixel level based on the free-form text with various open-set vocabularies.
Unlike traditional single-modality image segmentation, which assigns predefined labels to all pixels, RIS requires cross-modal understanding to locate and segment the referent at the pixel level, guided by free-form text based on diverse open-set vocabularies.
Referring remote sensing image segmentation (RRSIS) extends this task to the remote sensing domain, advancing vision-and-language developments in complex aerial scenarios.
It attracts increasing attention and is crucial for decision-making with rich human-friendly text-prompts in real applications, including urban infrastructure management \cite{duan2016towardsurban}, post-disaster studies \cite{kalluri2024robustdisaster}, and land use/cover change survey \cite{liu2024remoteclipchange}.
Given the diverse spatial scales of ground targets and complex backgrounds in high-resolution aerial images, fully leveraging cross-modal interactions to achieve accurate pixel-level predictions remains highly challenging.

\begin{figure*}[t]
  \centering
  \begin{subfigure}[t]{0.8\linewidth}
  \includegraphics[width=1\linewidth]{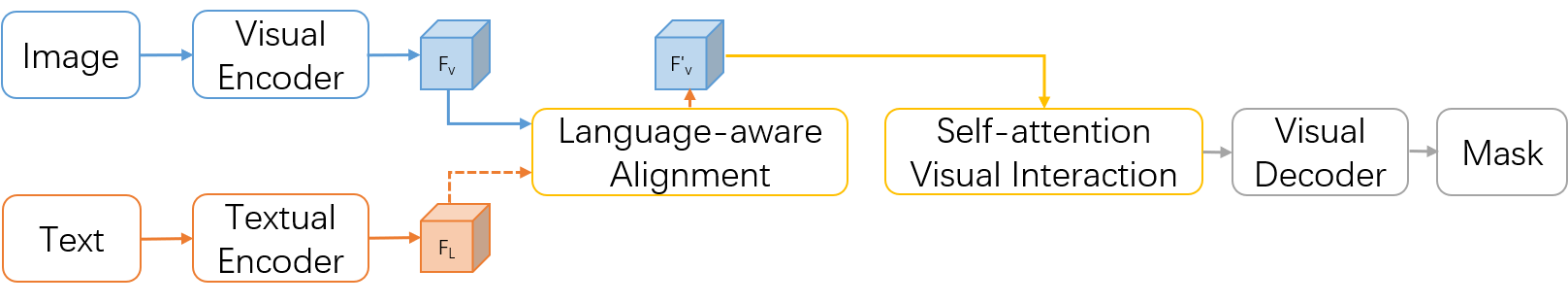}
  \subcaption{Language-aware alignment with a self-attention visual interaction module.}
  \label{fig: a}
  \end{subfigure}
  % \hspace{2mm}
  % \hfill
  \begin{subfigure}[t]{0.8\linewidth}
  \includegraphics[width=1\linewidth]{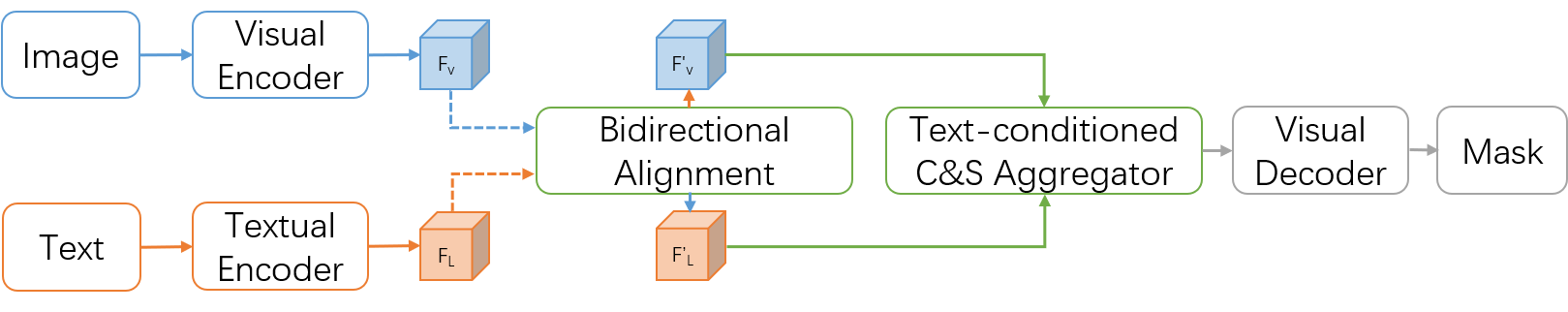}
  \subcaption{Our bidirectional alignment with a text-conditioned aggregator.}
  \label{fig: b}
  \end{subfigure}

  \begin{subfigure}[t]{0.8\linewidth}
  \subcaption*{\textbf{$Expression$:} The red baseball diamond on the right}
  \end{subfigure}
 
  \begin{subfigure}[t]{0.19\linewidth}
  \includegraphics[width=0.95\linewidth]{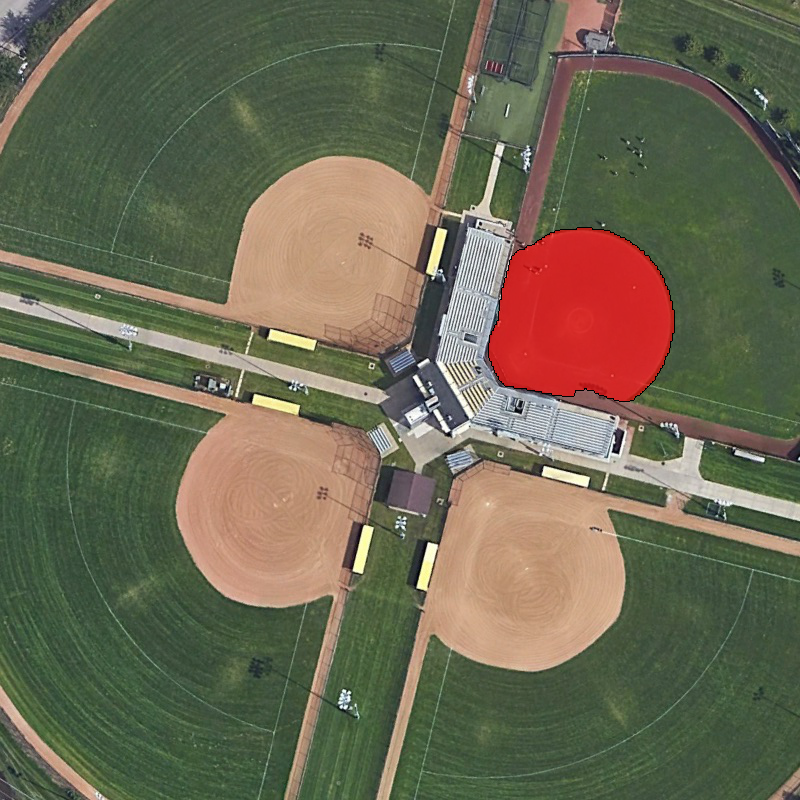}
  \subcaption{}
  \label{fig: c}
  \end{subfigure}
  \begin{subfigure}[t]{0.19\linewidth}
  \includegraphics[width=0.95\linewidth]{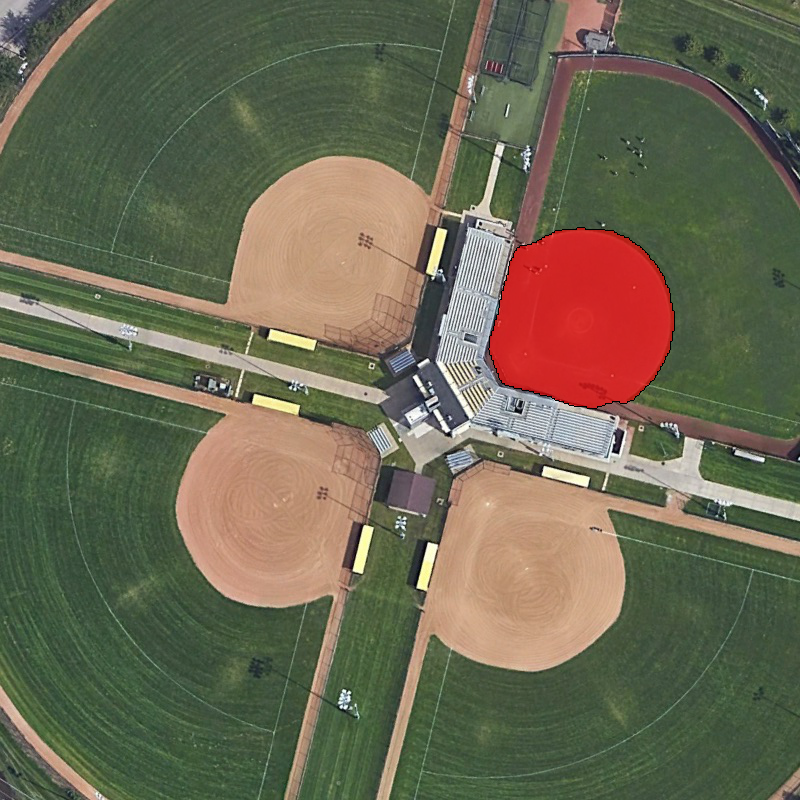}
  \subcaption{}
  \label{fig: d}
  \end{subfigure}
  \begin{subfigure}[t]{0.19\linewidth}
  \includegraphics[width=0.95\linewidth]{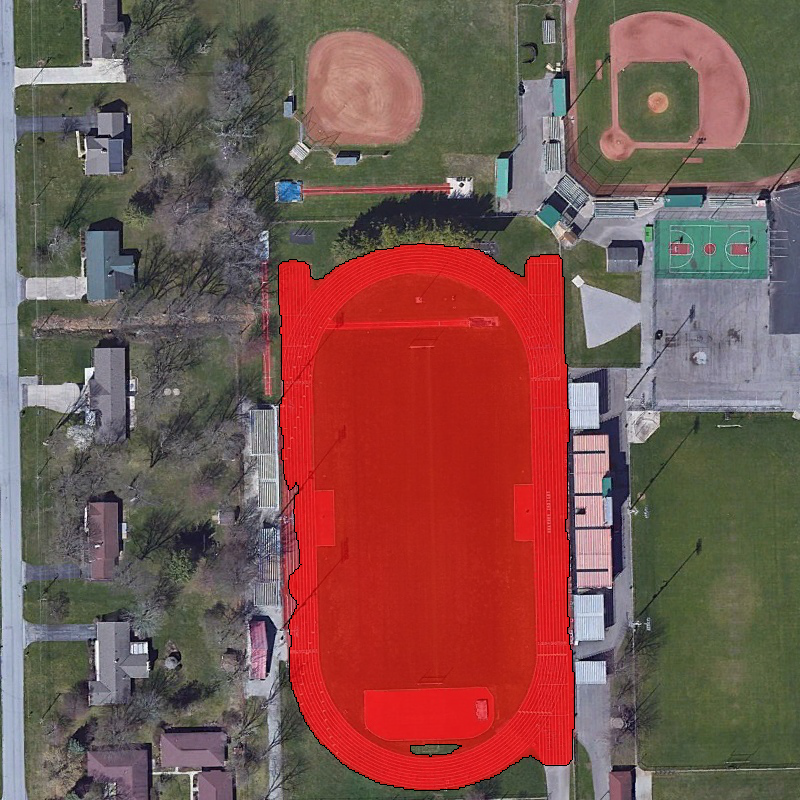}
  \subcaption{}
  \label{fig: e}
  \end{subfigure}
  \begin{subfigure}[t]{0.19\linewidth}
  \includegraphics[width=0.95\linewidth]{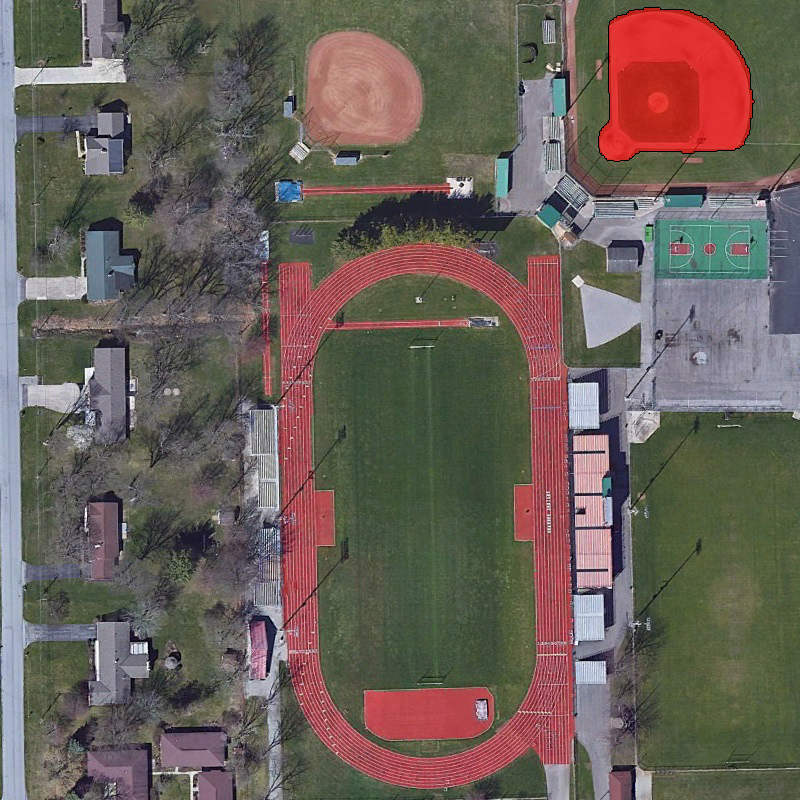}
  \subcaption{}
  \label{fig: f}
  \end{subfigure}
  
  % \begin{subfigure}[t]{0.8\linewidth}
  
  \vspace{-2mm}
  
  % \subcaption{Two examples sharing the same expression and the predictions made by (a) and (b), respectively.}
  % \label{fig: c}
  % \end{subfigure}
  
  \caption{Illustration of different methods for RRSIS and their corresponding results on two examples from the RRSIS-D \cite{liu2024rmsin} dataset.
  In (a), the previous methods utilize language-aware transformers to only update visual features, while our method selectively refines both visual and linguistic features with vision-to-language and language-to-vision flows in (b).
  To distinguish the inner guidance in the cross-modal alignment modules, we represent the directional flows with dashed arrows in orange and blue.
  We further present two examples sharing the same expression: (c) and (e) are results obtained using (a), while (d) and (f) are results predicted by our method.
  }
  \label{fig: illustration}
  \vspace{-3mm}
\end{figure*}

Existing approaches to referring segmentation construct their frameworks by integrating per-pixel classification with multi-modal feature fusion.
Inspired by mainstream RIS methods \cite{huang2020cmpc, liu2021cmpc+, yang2022lavt}, these techniques adopt a straightforward representation-fusion-segmentation pipeline and utilize diverse feature fusion architectures to facilitate cross-modal interaction
For example, LGCE \cite{yuan2024rrsis} improved the original pixel-word alignment from LAVT \cite{yang2022lavt} with a language-guided cross-scale fusion module to combine the shallow- and deep-layer visual features.
RMSIN \cite{liu2024rmsin} developed the language-guided fashion with a visual gate and further considered scales and orientations by leveraging inter- and cross-scale fusion and rotated convolutions.

Although these methods have achieved impressive segmentation performance for RRSIS, they still face several limitations in aerial scenarios.
% First, the methods \cite{yuan2024rrsis, liu2024rmsin} employing single direction flow (\ie~language-guided fusion) equally treat each token from the given text and fix them during the feature refinement in the encoder, which fails to capture the content-wise visual perception with linguistic features from the complex ground.
First, methods utilizing a single-directional flow (\emph{i.e.,}~language-guided fusion) treat all tokens in the given text equally and keep them fixed during feature refinement in the encoder.
They fail to adaptively capture content-aware visual perception guided by linguistic features, particularly in the context of complex ground environments.
As shown in Fig.~\ref{fig: c}, previous methods can generate a reasonable mask for the first example by focusing solely on the attribute ``red".
However, they struggle to differentiate between multiple objects of the same or different categories sharing the same color as shown in Fig.~\ref{fig: e}.
These results highlight the need for textual focus to adapt dynamically to varying visual content during cross-modal interaction.
% but fail to distinguish multiple objects with same and different categories in the same color for the second example.
% The results indicate that the textual focuses should also vary with different visual content during cross-modal interaction.
% the distinct visual region in the red circle indicates that the specific token (`red') highly contributes to the cross-modal interaction, which cannot be highlighted in such a single-modality flow.
Second, although some works \cite{liu2024rmsin, yuan2024rrsis, hu2023beyondscale, lei2024exploringarxiv} try to merge cross-scale visual information, they directly concatenate the hierarchical visual features and leverage standard self-processing operations (\emph{e.g.,}~self-attention \cite{vaswani2017attention}), before decoding the final predictions (as shown in Fig.~\ref{fig: a}).
However, the spatial priors derived from both visual and linguistic inputs impose greater demands on the representation of global context and local details.
Overall, existing works neglect the complementary vision-to-language guidance in cross-modal interaction and fail to preserve spatial relationships between multi-modal features during cross-scale information exchange.
These limitations lead to suboptimal segmentation performance, particularly in complex semantic scenarios.

In this paper, we propose an effective and domain-specific RRSIS framework to address the aforementioned limitations.
Instead of concentrating solely on refining visual features during cross-modal interaction, we adopt a bidirectional alignment scheme (as shown in Fig.~\ref{fig: b}) to selectively represent the visual and linguistic features and improve the crucial information exchange for the multi-modal segmentation task.
Specifically, we propose a Scale-wise Bidirectional Alignment Network (SBANet) consisting of bidirectional alignment modules and a text-conditioned channel and spatial aggregator.
The bidirectional alignment module aims to refine the multi-modal features from both vision-to-language and language-to-vision directions, which representatively update the features in the encoder according to the visual content and textual tokens.
Different from the previous methods that keep linguistic features fixed during cross-modal interaction, we first propose learnable query tokens to sparsely and effectively represent the visual context and then use them to update related linguistic features of the crucial textual tokens.
Besides, we enhance the alignment module with a dynamic feature selection block to capture global context and local details during the visual feature refinement at each scale level in the hierarchical encoder.
Furthermore, leveraging multi-level features to decode the final mask is essential for accurate high-resolution pixel-wise predictions.
% utilizing multi-level features to decode the final mask is also crucial for the high-resolution pixel-wise predictions.
To this end, we propose a text-conditioned aggregator that incorporates both channel- and spatial-wise attention mechanisms to improve query guidance during cross-scale reasoning.
% Therefore, we propose a text-conditioned aggregator with both channel- and spatial- wise attentions to enhance the query guidance within the cross-scale reasoning.
Through the proposed modules, our SBANet effectively represents the multi-modal features and achieves the bidirectional guidance, thereby enhancing its capability to accurately discern the ground targets within and across different scales.
The \textbf{main contributions} of this paper are summarized as follows:
\begin{itemize}
    \item We propose a Scale-wise Bidirectional Alignment Network (SBANet) for RRSIS that leverages bidirectional alignment modules to effectively achieve cross-modal interaction and refine both visual and linguistic features from language-to-vision and vision-to-language directions.
    \item We propose learnable query tokens to help refine linguistic features by a query-text token alignment module, and a dynamic feature selection block to capture global context and local details, updating the visual features within each hierarchy.
    \item In addition, we design a text-conditioned channel and spatial aggregator to facilitate cross-scale information exchange with both spatial and channel focuses before decoding the final results.
    \item Extensive experiments on two challenging RRSIS benchmarks demonstrate that our proposed method achieves superior segmentation results in comparison to state-of-the-art methods.
\end{itemize}

\section{Related Work}
\label{related work}
In the last decade, we have witnessed impressive advancements in developing deep learning methods for locating targets with both visual and textual inputs in computer vision and remote sensing communities.
In this section, we review the most relevant works to our RRSIS work proposed in this paper.

\subsection{Referring Image Segmentation}
Referring image segmentation (RIS) aims to segment target objects in images according to given natural language expressions.
Normally, RIS involves creating separate or joint representations for the multi-modal inputs, followed by a feature fusion stage.
% requires separate or joint representations for the multi-modal inputs and a following feature fusion stage.
Early works \cite{nagaraja2016modelingearly2, hu2016segmentationearly3, li2018referringearly1} employ standard backbones (\emph{e.g.,}~convolution networks \cite{simonyan2014vgg, he2016deepresnet} and recurrent neural networks \cite{hochreiter1997lstm}) to extract visual and linguistic features, respectively, and fuse them with a simple concatenation step for the final predictions.
Subsequent methods \cite{shi2018keylevel, yang2020propagatinglevel, shi2022spatiallevel} try to analyze the structure of natural language expressions across multiple levels, including word-level, sentence-level, and group-level encodings.
For example, some group-based methods decompose expressions into different groups through explicit (\emph{e.g.,}~concept and relationship \cite{yang2019crossexpl, hui2020linguisticexpl}) or implicit operations (\emph{e.g.,}~attention mechanisms \cite{fu2019dualimc, wu2020phrasecutimc, yang2021bottomimc}).
Although these textual encodings from different levels provide various linguistic understanding, they only involve single-modality representation but fail to sufficiently interact with the other modality.
To overcome the limitation, some works introduce various vision-language alignment strategies for improving the interaction between two modalities, including methods based on progressive refinement \cite{liu2017recurrentrefi, chen2019seerefi} and dynamic filters \cite{margffoy2018dynamicfil, chen2019referringsub, jing2021locatesub, liu2023caris}.
The recent emergence of Transformers \cite{vaswani2017attention} further pushes the study of RIS, which provides robust and powerful fusion capability for multi-modal integration.
For instance, CMSA \cite{ye2019csma} proposed cross-modal self-attention modules to capture the long-range dependencies between visual and linguistic features.
Subsequently, VLT \cite{ding2021visionvlt} built a Transformer framework equipped with proposed query generation modules that represent the given language from different aspects for enriching textual comprehension.
To further improve cross-modal integration, LAVT \cite{yang2022lavt} proposed a robust hierarchical baseline that employs early fusion through language-aware attentions.
Building upon this, GRES \cite{liu2023gres} explicitly interacted with different visual regions and textual tokens to analyze their dependencies, thereby improving segmentation performance.
However, these existing RIS methods designed for natural images are limited when dealing with diverse scales of ground targets in complex environment from high-resolution aerial images, resulting in suboptimal performance in aerial scenarios.

\subsection{Referring Remote Sensing Image Segmentation}
Referring remote sensing image segmentation (RRSIS) requires to extract pixel-wise ground target masks from aerial images according to specific natural language expressions.
The exploration of RRSIS remains relatively scarce, and only a few works have been developed to fill in the domain gap between natural images and aerial images.
LGCE \cite{yuan2024rrsis} first introduced a remote sensing dataset (\emph{i.e.,}~RefSegRS) designed for RRSIS and improved LAVT \cite{yang2022lavt} with shallow- and deep-layer features fusion.
Given the limited number and diversity of samples in the RefSegRS dataset, Liu et al. \cite{liu2024rmsin} constructed a larger dataset, RRSIS-D, and benchmarked mainstream RIS methods on this new dataset.
% Since the quantity of the samples in the RefSegRS dataset is quite small, RMSIN \cite{liu2024rmsin} further built a larger dataset termed RRSIS-D 
In addition, it proposed a rotated multi-scale interaction network with intra- and cross-scale interaction modules built on the top of LAVT.
Pan et al. \cite{pan2024mmrrsis} analyzed the current implicit optimization paradigm and designed an explicit affinity alignment, incorporating a new loss function.
% equipped with a new loss function.
However, these methods primarily focus on the language guidance when processing hierarchical visual features in the encoder but fail to capture the correspondence between the visual context to key textual tokens.
The complex visual background in aerial scenarios poses more challenges to specific linguistic representation.
Very recently, FIANet \cite{lei2024exploringarxiv} proposed a fine-grained image-text alignment module with object-positional enhancement, along with a text-aware self-attention module, to process concatenated multi-scale features.
CroBIM \cite{dong2024crossarxiv} introduced a context-aware prompt modulation module to process the post-fusion of multi-scale visual features and linguistic features, and a mutual-interaction decoder to enhance segmentation performance.
Although these methods improve the refinement of visual features through various strategies, they still treat linguistic features as fixed from the original encoder (\emph{e.g.,}~BERT \cite{devlin2018bert}) and apply equal treatment to all textual tokens during alignment.
In addition, simply incorporating vanilla self-attentions on multi-scale features neglects the spatial information inherent in both visual context (\emph{e.g.,}~relative locations) and textual tokens (\emph{e.g.,}~specific descriptions).
In contrast, our proposed SBANet updates both visual and linguistic features during cross-modal alignment and employs a text-conditioned channel and spatial aggregator to enhance multi-scale information exchange.

\subsection{Visual Grounding for Aerial Images}
Similar to RRSIS, remote sensing visual grounding (RSVG) is another active field for vision-and-language research in remote sensing community \cite{zhao2021highic, zhan2023rsvg, li2024hrvqa}.
It also aims to locate referent targets within aerial images according to given natural language expressions, but differs in only requiring object-level predictions (\emph{e.g.,}~bounding boxes) in comparison to RRSIS.
GeoVG \cite{sun2022visualgeovg} first introduced RSVG and proposed a one-stage framework by leveraging a geospatial relationship graph to represent linguistic features.
Subsequent works (\emph{e.g.,}~MGVLF \cite{zhan2023rsvg} and VSMR \cite{ding2024visualrsvgnew}) further explored the special visual context from aerial images with multi-scale and multi-granularity fusion.
Li et al. \cite{li2024languagersvgnew2} proposed a language-aware progressive visual attention network via dynamically generated multi-scale weights and biases for key information extraction and the suppression of irrelevant regions.
However, these methods cannot be directly applied to RRSIS due to differences in the primary focus of the network designs.
% However, these methods cannot be directly employed for RRSIS because the main focus of network designs is different.
While RSVG methods process linguistic features to locate referent objects in a relatively coarse manner, subsequent strategies like progressive refinement help generate object-level bounding boxes.
% Since RSVG methods process linguistic features to locate the referent objects in a relatively coarse way, some following strategies such as the progressive refinement help to generate object-level bounding boxes.
Differently, RRSIS requires precise pixel-wise predictions conditioned on specific natural language expressions.
This necessitates a more involved approach, where not only the cross-modal interaction at each scale is crucial, but also the processing of these interactions prior to decoding the final masks.

\section{Method}
\label{sec: methodology}
In this section, we first introduce the overview of the proposed SBANet and our adopted baseline model in Sec.~\ref{sec: overview}. Then, we elaborate on the proposed bidirectional alignment module in Sec.~\ref{sec: bam}. Finally, we detail the text-conditioned channel and spatial aggregator in Sec.~\ref{sec: aggregator}.
%------------------------------------------------------------------------
\begin{figure*}[t]
  \centering
  \includegraphics[width=0.85\linewidth]{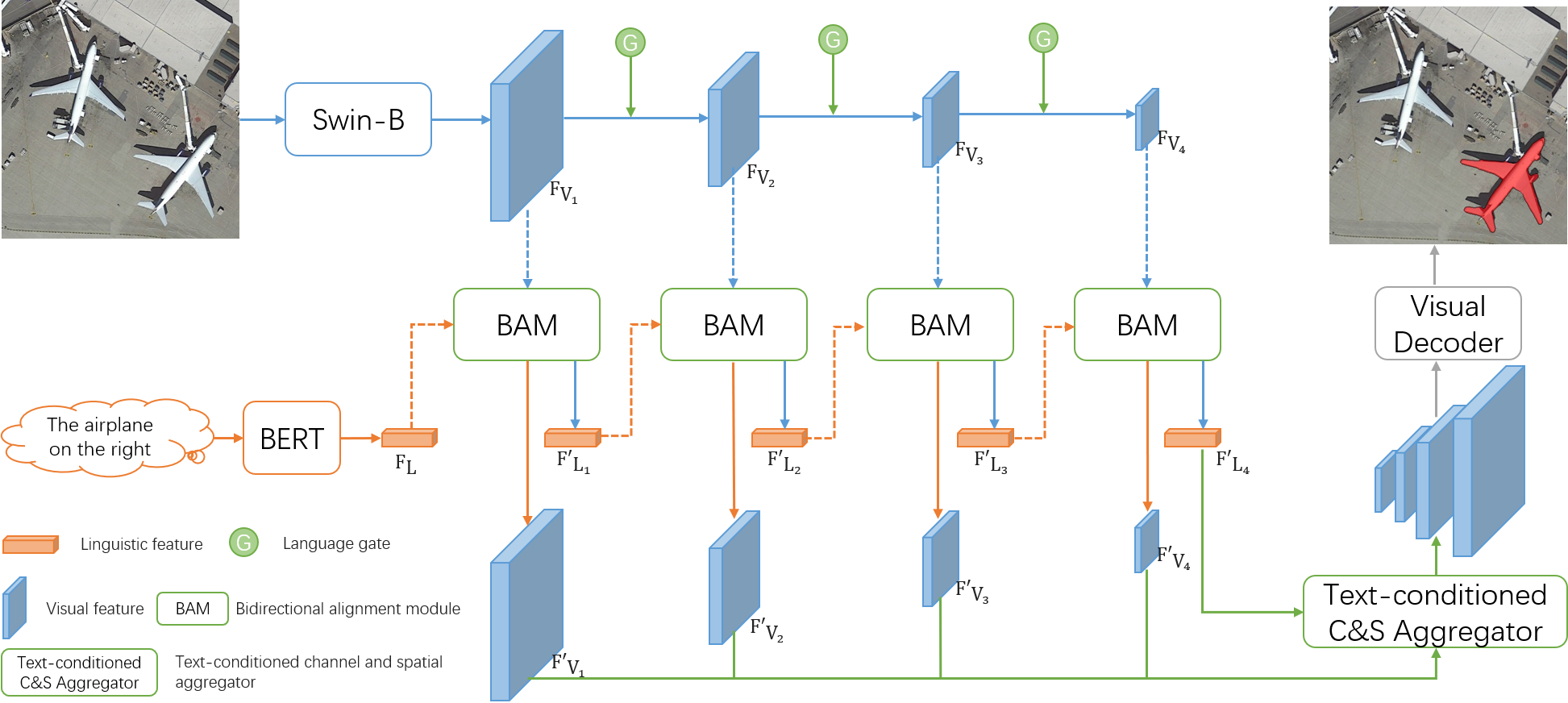}
  \caption{The overview of SBANet framework.
  We utilize the base Swin Transformer \cite{liu2021swin} and BERT \cite{devlin2018bert} as visual and textual encoders for extracting visual and linguistic features, respectively.
  At the first stage, the visual features $F_{V_1}$ and original linguistic features $F_{L}$ are fed into the proposed bidirectional alignment module (BAM) to obtain updated results $F'_{V_1}$ and $F'_{L_1}$.
  We illustrate the different directional flows using dashed arrows in orange and blue.
  Then $F'_{V_i}$ are employed through a language gate (on the upper part) to generate $F'_{V_{i+1}}$ at the next stage while $F'_{L_1}$ are used as input for the next BAM.
  With the updated visual and linguistic features together, we introduce a text-conditioned channel and spatial aggregator to enhance cross-scale information exchange before decoding the final mask.
  For brevity, we do not show the concatenation for the decoder.
  }
  \label{fig: overview}
  \vspace{-2mm}
\end{figure*}

\subsection{Overview}
\label{sec: overview}
The overall architecture of our proposed SBANet is illustrated in Fig.~\ref{fig: overview}.
An aerial image $I\in \mathbb{R}^{H \times W \times C}$ and a natural language expression $T = \{w_t\}, t \in \{0, ..., {N-1}\}$ are fed into the network as the input, where $H$, $W$, and $C$ denote the height, width, and dimensionality of the image, and $N$ represents the number of words $w_t$ in the input text.
For the linguistic feature extraction, we utilize a pre-trained powerful language encoder (\emph{e.g.,}~BERT \cite{devlin2018bert}) for processing the input expression by converting the textual tokens to robust linguistic features $F_L\in \mathbb{R}^{l \times d}$, where $l$ and $d$ denote the maximum length of tokens and the dimensionality of the linguistic features, respectively.
For the visual feature extraction, we follow mainstream RIS methods \cite{yang2022lavt, liu2023caris, liu2023gres} by employing a pre-trained hierarchical encoder (\emph{e.g.,}~Swin Transformer \cite{liu2021swin}) to generate multi-scale visual features $F_V\in \mathbb{R}^{h_i \times w_i \times c_i}$, where $i \in \{1, 2, 3, 4\}$ represents the hierarchical stage.
We take the LAVT \cite{yang2022lavt} model as our baseline that includes the same encoders for the extraction of multi-modal inputs and fuses the linguistic features $F_L$ with current visual features $F_{V_{i-1}}$ to generate $F_{V_i}$ for the next stage.
Note that we keep the same settings of the architecture in LAVT as these parts are not our focus in this paper.
To selectively represent the visual and linguistic features in cross-modal fusion and improve the interaction, we replace the original pixel-word attention module (a brief explanation can be found in Sec.~\ref{sec: revisit}) in LAVT with our proposed bidirectional alignment module.
Furthermore, we introduce a text-conditioned aggregator to mitigate the gap between the encoder and decoder accompanied with linguistic focus, and enhance cross-scale information exchange with channel and spatial attentions.

%---------------------------------------------------------------------
\subsection{Bidirectional Alignment Module}
\label{sec: bam}
\begin{figure}[t]
  \centering
  \includegraphics[width=1\linewidth]{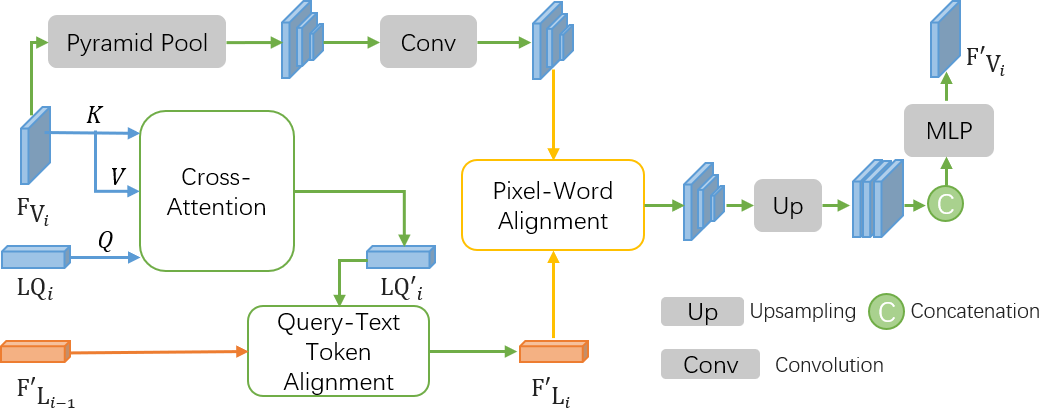}
  \caption{The architecture of the proposed BAM.
  The learnable query tokens enable sparsely represent the visual context with position embedding (not shown here for a clear presentation), and they are then used to guide the update of linguistic features through a query-text token alignment module.
  A dynamic feature selection block processes pyramid visual features for a selective pixel-word alignment with $F'_{L_i}$, subsequently obtaining updated visual features $F'_{V_i}$ with upsampling and MLP.
  }
  \label{fig: bam}
  \vspace{-5mm}
\end{figure}

To effectively capture the most crucial information from both visual and linguistic features, selective representation in the multi-scale encoder is essential for cross-modal interaction in high-resolution aerial scenarios.
We propose a bidirectional alignment module (BAM) to improve the interaction, which consists of learnable query tokens for sparse visual representation, query-text token alignment to update linguistic features, and a dynamic feature selection block for the refinement of visual features, as shown in Fig.~\ref{fig: bam}.
The details about each component are introduced in the following subsections.

\subsubsection{Revisit of Pixel-Word Alignment Module}
\label{sec: revisit}
Before delving into the detailed design of BAM, we briefly revisit the pixel-word alignment module (PWAM) proposed in LAVT \cite{yang2022lavt}, which coarsely present the popular cross-modal interaction in RIS.
PWAM generally includes two steps for multi-modal feature fusion.
First, it integrates the pre-trained linguistic features $F_L$ across the word dimension with visual features $F_{V_i}$ at each spatial position, resulting in a position-specific, sentence-level feature vector.
Specifically, the features go through size-related projection functions with 1 $\times$ 1 convolution and normalization layers.
Then it involves various rolling, unrolling, and transposing operations to obtain features in proper shapes for cross-attentions.
The core scaled dot-product attention \cite{vaswani2017attention} takes the query from visual features and the keys and values from linguistic features to obtain cross-attended results.
Second, the module combines the reshaped results with the original visual features via an element-wise multiplication followed by a multi-modal projection including 1 $\times$ 1 convolution and ReLU \cite{nair2010rectifiedrelu} nonlinearity.
The output from PWAM is further processed as a language gate to compromise multi-modal residuals, avoiding disruption of pure vision initializations from the encoder.

\subsubsection{Linguistic Feature Update with Learnable Query Tokens}
The aforementioned PWAM processes linguistic features as the language guidance to update visual focuses.
However, as discussed in Section~\ref{sec: intro}, the linguistic features should also be updated for varying multi-scale visual context.
Different from vanilla cross-attentions adopted in CARIS \cite{liu2023caris}, we introduce a set of learnable query tokens to sparsely represent current visual context and then utilize the query tokens to achieve vision-to-language guidance.

Specifically, for each BAM at a hierarchical scale, we randomly initialize \textit{M} learnable query tokens $LQ_i\in \mathbb{R}^{M \times c_q}$ to sparsely highlight crucial visual content, where $c_q$ denotes the dimensionality of the query tokens.
These query tokens capture the most representative information from images at the current scale, increasing the likelihood of effective interaction with referring expressions.
To achieve that, we take the queries from the initialized learnable query tokens $LQ_i$ and the keys and values from visual features $F_{V_i}$ for a cross-attention block.
Inspired by the application of relative positioning in image recognition \cite{cheng2021perpe, wu2022seqformerpe}, we also add a randomly initialized position embedding accompanied with the query tokens.
The linear projection and normalization layer are employed to obtain the corresponding matrices.
Then a scaled dot-product attention computes the updated query tokens by,
% \begin{equation}
%     Q_{LQ_i} = DConv_C(F_{V_i}), K_{C} = DConv_C(F_{C}), V_{C} = DConv_C(F_{C}),
% \end{equation}
\begin{equation}
     f_{LQ}(Q_{LQ_i},K_{F_{V_i}},V_{F_{V_i}}) = {\rm Softmax}(\frac{Q_{LQ_i}K_{F_{V_i}}^T}{\sqrt{c_i}})V_{F_{V_i}},
\end{equation}
where $Q_{LQ_i}$, $K_{F_{V_i}}$, and $V_{F_{V_i}}$ represent queries, keys and values, respectively.
We further adopt a multi-layer perceptron (MLP) and layer normalization (LN) \cite{ba2016layerln} on the output of this function to collect the position-aware learnable query tokens $LQ_i'$.

Subsequently, the update of linguistic features is achieved with the help of the sparse representative visual context rather than the self-correspondences among words or cross-correlations with entire visual features.
Specifically, the updated learnable query tokens $LQ_i'\in \mathbb{R}^{M \times c_q}$ and linguistic features $F'_{L_{i-1}}\in \mathbb{R}^{l \times d}$ from the last stage ($F_L$ for the first BAM) are fed into a query-text token alignment module.
First, it captures the visual information most relevant to the natural expression at each token position.
We project $LQ_i'$ into a common feature space by leveraging a 1 $\times$ 1 convolution and a GeLU \cite{hendrycks2016gaussiangelu} layer, and apply the similar operation to $F'_{L_{i-1}}$.
The crucial query feature maps $R_i$ are computed by,
\begin{equation}
     R_{iqt} = {\rm Softmax}(\frac{w_{iq}(F'_{L_{i-1}})w_{ik}(LQ_{i}'^T)}{\sqrt{c_q}})w_{iv}(LQ_{i}'),
\end{equation}
\begin{equation}
     R_i = w_{iqt}(Reshape(R_{iqt})),
\end{equation}
where $w_{iq}$, $w_{ik}$, $w_{iv}$, and $w_{iqt}$ represent query, key, value, and query-text projection functions, repetitively, and \textit{Reshape} denotes the operation used to obtain feature maps with the same shape as $F'_{L_{i-1}}$.
The query-text projection is implemented by a 1 $\times$ 1 convolution with normalization.
Second, we utilize element-wise multiplication on $R_i$ and $F'_{L_{i-1}}$ to obtain the updated linguistic features $F_{L_i}'$ conditioned on the current $i_{th}$ visual context by,
\begin{equation}
     F_{L_i}' = w_{if}(w_{il}(F'_{L_{i-1}}) \odot R_i),
\end{equation}
where $w_{il}$ and $w_{if}$ represent linguistic and final projections, repetitively, and $\odot$ denotes the element-wise multiplication.
The projections also include a 1 $\times$ 1 convolution followed by ReLU nonlinearity as PWAM.
Till now, we update the linguistic features based on the crucial visual information that selectively present the textual focuses for the following visual feature refinement stage.

\subsubsection{Visual Feature Update with Dynamic Feature Selection Block}
On top of the updated linguistic features, we produce referring visual features by a dynamic feature selection block for the language-to-vision guidance.
To effectively align visual features with the selective linguistic features, we employ a pyramid visual representation structure to globally capture the visual context and find the most relevant image regions to the referents.
Specifically, the visual features $F_{V_i}$ at each scale are first fed into a pyramid group (\emph{e.g.,}~\{1, 2, 3, 6\}, representing different sub-scales), which includes different sizes of pooling windows for collecting dynamic visual representation.
Then the visual features are further processed by three 1 $\times$ 1 convolutions with normalization for refining global information at different sub-scales.
We integrate the obtained pyramid visual features $F_{V_i}^p$ with selective linguistic features $F_{L_i}'$ by following PWAM.
After that, we obtain dynamic visual features across different sub-scales guided by textual focuses.
These features are further combined through separate upsampling operations via bilinear interpolation followed by a channel-wise concatenation.
To reduce the dimensionality of the concatenated features and improve valid representation, we employ a two-layer MLP with LN for obtaining the final visual features as the aligned results.
The dynamic feature selection block is mathematically described as follows,
\begin{equation}
     F_{V_i}^p = \text{Conv}(\text{PyramidPool}_{group}(F_{V_i})),
\end{equation}
\begin{equation}
     F_{cross}^p = \text{PWAM}_{group}(F_{L_i}', F_{V_i}^p),
\end{equation}
\begin{equation}
     F_{V_i}' = \text{MLP}(\text{Concat}(\text{Up}(F_{cross}^p))),
\end{equation}
where $\text{PyramidPool}_{group}$, $\text{PWAM}_{group}$, $\text{Concat}$, and $\text{Up}$ denote pyramid pooling operation, pixel-word alignment \cite{yang2022lavt}, channel-wise concatenation, and upsampling operation, respectively.
For brevity, the adopted normalization operations are omitted.
Similar to the operation performed in LAVT, we utilize a language gate acting as residuals to allow an adaptive amount of guidance flowing to the next stage.

In summary, our proposed BAM not only updates visual features through sparse and dynamic feature representations aligned with textual guidance but also refines linguistic features through query-text token alignment.
These components selectively extract the most relevant information to the visual context and textual focuses, thereby improving cross-modal alignment for referring segmentation.

%---------------------------------------------------------------------
\subsection{Text-Conditioned Channel And Spatial Aggregator}
\label{sec: aggregator}
\begin{figure}[t]
  \centering
  \includegraphics[width=1\linewidth]{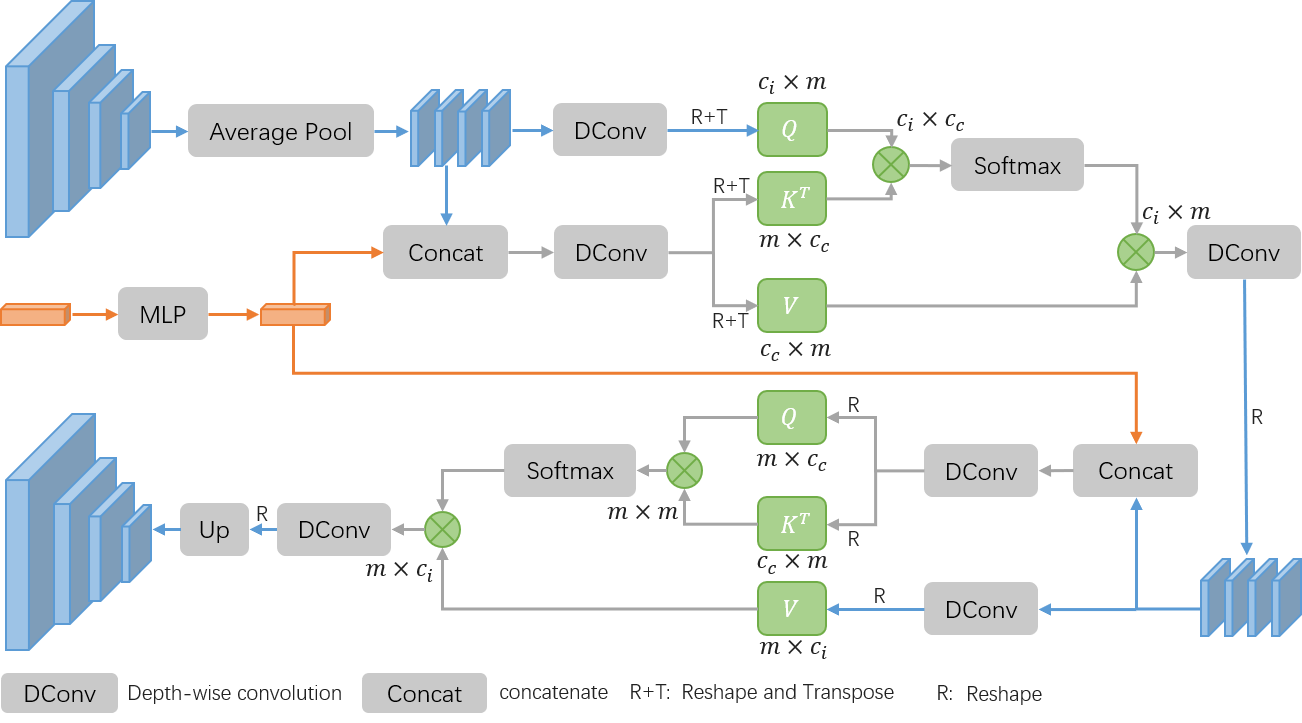}
  \caption{The architecture of the proposed text-conditioned channel and spatial aggregator (TSCA).
  The shapes of queries, keys and values are also present, where $m$, $c_i$, and $c_c$ denote the number of patches, the number of channels at each scale and the number of all concatenated features, respectively.
  The updated linguistic feature are incorporated with multi-scale visual features through channel and spatial attentions before decoding the final mask.
  It produces enhanced multi-scale representations and connects between the multi-modal encoder and visual decoder.
  }
  \label{fig: tcsa}
  \vspace{-5mm}
\end{figure}

When encountering scale discrepancy among the updated visual features from the encoder, directly integrating them with the decoder may lead to spatial disconnectedness in pixel-wise predictions.
Alternative solutions such as the cross-scale interaction proposed in \cite{liu2024rmsin} employ vanilla self-attentions to process the concatenated visual features before the decoder.
However, different from standard semantic segmentation, RIS requires cross-modal understanding of both visual content and linguistic expressions.
Treating the concatenated visual features equally without guidance from the textual side may lead to suboptimal segmentation performance
% Equally treating the concatenated visual features without the guidance from the textual part could result in suboptimal segmentation performance.
To address this issue, we propose a text-conditioned channel and spatial aggregator as shown in Fig.~\ref{fig: tcsa}.
First, we leverage the updated linguistic features from BAM to extract textual focuses through a recapping operation before integrating them with multi-scale visual features.
Then channel and spatial attentions extract the most crucial focuses for the referring context.

Specifically, we first utilize a two-layer MLP with LN to recap the linguistic features $F_{L_i}'$, and project them to the space of visual features for obtaining the textual guidance $F_{L_g}$.
For brevity, we only use the updated linguistic features $F_{L_4}'$ from the last BAM here for $F_{L_g}$.
Subsequently, we feed the updated multi-scale visual features $F_{V_i}'$ and the textual guidance $F_{L_g}$ to a sequential channel and spatial attention block.
For channel attentions, the visual features $F_{V_i}'$ first undergo a layer normalization operation.
Then they are concatenated with $F_{L_g}$ along the channel dimensionality to form $F_{C}$, following average pooling and reshaping.
We obtain keys and values from the concatenated results $F_{C}$ and queries from each scale $F_{V_i}'$ for the cross-attention processing.
Differently, we replace the original linear projections \cite{vaswani2017attention} with 1 $\times$ 1 depth-wise convolutions for capturing local details and reducing computational cost as proved by prior works \cite{guo2022cmt, lee2022mpvit}.
In addition, the projection operates on each channel and obtains the final results through a fusion step.
The projected queries $Q_{C_i} \in \mathbb{R}^{m \times c_i}$, keys $K_{C} \in \mathbb{R}^{m \times c_c}$, and values $V_{C} \in \mathbb{R}^{m \times c_c}$ are computed as follows,
\begin{equation}
    \begin{split}
        Q_{C_i} = DConv_C(F_{V_i}'),\\
        K_{C} = DConv_C(F_{C}),\\
        V_{C} = DConv_C(F_{C}),
    \end{split}
    % Q_{C_i} = DConv_C(F_{V_i}'),  K_{C} = DConv_C(F_{C}), V_{C} = DConv_C(F_{C}),
\end{equation}
where $m$, $c_i$, and $c_c$ denote the number of patches, the number of channels at each scale and the number of all concatenated features.
Thus, the channel attentions are obtained through the transpose of the items by,
\begin{equation}
     f_{C}(Q_{C_i},K_C,V_C) = {\rm Softmax}(\frac{Q^T_{C_i}K_C}{\sqrt{c_c}})V_C^T,
\end{equation}
Then we employ depth-wise convolutions and reshape the outputs before fed into spatial attentions.
For spatial attentions, we also apply the layer normalization and concatenation to the reshaped features.
Differently, we take queries and keys from the concatenated result and values from each scale.
Then the projected queries $Q_{S} \in \mathbb{R}^{m \times c_c}$, keys $K_{S} \in \mathbb{R}^{m \times c_c}$, and values $V_{S_i} \in \mathbb{R}^{m \times c_i}$ are computed as follows,
\begin{equation}
\begin{split}
    Q_{S} = DConv_S(F_{C}),\\
    K_{S} = DConv_S(F_{C}),\\
    V_{S_i} = DConv_S(F_{V_i}'),
\end{split}
\end{equation}
The spatial attentions do not require the transpose operation but need to put the number of heads $n_h$ into consideration by,
\begin{equation}
     f_{S}(Q_{S},K_S,V_{S_i}) = {\rm Softmax}(\frac{Q_SK_S^T}{\sqrt{c_c/n_h}})V_{S_i},
\end{equation}
Then we reshape the outputs and apply depth-wise convolutions before decoding the final masks.
Note that the two kinds of aggregators have different objectives for cross-scale information exchange.
Channel attentions process channel-wise focuses by integrating all spatial positions along given channels while spatial attentions capture spatial context by leveraging spatial inter-dependencies between any two channels along specific positions.
Moreover, cross-scale information exchange captures cross-correlations between low- and high-level semantics and highlights the referent-related features, thereby suppressing irrelevant features and enhancing segmentation performance.

%---------------------------------------------------------------------

\section{Experiments}
In this section, we discuss the experiments conducted to assess our proposed SBANet for RRSIS.
We first introduce the experiment setup, including datasets, implementation details and specific metrics for quantitative evaluations in Sec.~\ref{sec: setup}.
Then we perform the quantitative comparison with state-of-the-art methods on the evaluation datasets in Sec.~\ref{sec: comparison}.
To further analyze the effectiveness of our proposed method, we ablate the core designs in Sec.~\ref{sec: ablation}.
Finally, we present some qualitative examples in Sec.~\ref{sec: qualitative} to visually demonstrate the superiority of our method in aerial scenarios.

\subsection{Experiment Setup}
\label{sec: setup}

\subsubsection{Datasets}
\label{sec: datasets}
To evaluate the proposed method, we conducted extensive experiments on two available RRSIS datasets, RRSIS-D \cite{liu2024rmsin} and RefSegRS \cite{yuan2024rrsis}.
\begin{itemize}
  \item \textbf{RRSIS-D.} The dataset comprises a collection of 17,402 images with spatial resolution ranging from 0.5 to 30 meters.
  It divides the examples into three subsets, following a distribution of 12,181, 1,740, and 3,481, respectively.
  The dataset offers 20 category targets with seven potential attributes, which is quantitatively large and semantically rich for RRSIS.
  % \item \textbf{RefSegRS.} The dataset contains three subsets with image-expression-mask triplets for the referring segmentation task (training: 2,172 examples, validation: 431 examples, testing: 1,817 examples).
  % The examples cover 14 category ground objects described with five attributes to indicate the referents.
  % All images from this dataset are sized at 512 $\times$ 512 pixels, and the spatial resolution is 0.13 meters.
  % \item \textbf{RRSIS-D.} The dataset includes more examples than RefSegRS, which comprises a collection of 17,402 images with spatial resolution ranging from 0.5 to 30 meters.
  % It also divides the examples into three subsets, following a distribution of 12,181, 1,740, and 3,481, respectively.
  % Compared to RefSegRS, it offers 20 category targets with seven potential attributes, which is quantitatively larger and semantically richer for RRSIS.
  \item \textbf{RefSegRS.} The dataset contains three subsets with image-expression-mask triplets for RRSIS (training: 2,172 examples, validation: 431 examples, testing: 1,817 examples).
  Compared to RRSIS-D, the examples only cover 14 category ground objects described with five attributes to indicate the referents.
  All images from this dataset are sized at 512 $\times$ 512 pixels, and the spatial resolution is 0.13 meters.
\end{itemize}

\subsubsection{Implementation Details}
\label{sec: details}
We implemented our method in PyTorch \cite{paszke2019pytorch} and employed the pre-trained base BERT \cite{devlin2018bert} implementation from the HuggingFace's Transformer library \cite{wolf2020transformershugg]}.
% \footnote{https://huggingface.co/google-bert/bert-base-uncased}
For the visual encoder, we initialized the base Swin Transformer \cite{liu2021swin} with weights pretrained on ImageNet-22K \cite{deng2009imagenet}.
The default hyper-parameters in these encoders remained unchanged for the easy re-implementation.
The images with different sizes were resized to 448 $\times$ 448 pixels, and no data augmentation strategies (\emph{e.g.,}~rotation, flipping) were applied because of the use of location descriptions in the referring expressions.
% due to specific descriptions in expressions.
During the model training stage, we set the batch size to 12, and each model was trained for 40 epochs using \textit{AdamW} \cite{loshchilov2017decoupledadmw} with a weight decay of 0.01 and an initial learning rate of 0.0005.
Following our baseline LAVT \cite{yang2022lavt}, we utilized the cross-entropy loss \cite{mao2023cel} to train the model.
All experiments were conducted on two NVIDIA A40 GPUs.
% The above parameters are adopted in all the experiments unless specified otherwise.

\subsubsection{Evaluation Protocol}
For a fair comparison with previous methods \cite{liu2024rmsin, yang2022lavt, yuan2024rrsis, liu2023caris}, we adopted the same evaluation metrics, including mean Intersection over Union (mIoU), overall Intersection over Union (oIoU), and Precision at the 0.5, 0.7, and 0.9 threshold values (Pr@X).
Specifically, mIoU computes the average IoU between the predictions and ground truths across each test sample, which equally measures large and small objects from samples.
Differently, oIoU favors large objects as it computes the ratio of the total intersection area to the total union area of all test samples.
Moreover, Pr@X helps to assess the model performance of successfully predicted samples at different IoU levels.
Higher values for these evaluation metrics indicate better model performance.

%------------------------------------------------------------------------

\subsection{Quantitative Results and Comparison}
\label{sec: comparison}
We carried out a comprehensive comparison with state-of-the-art RIS methods on the RRSIS-D \cite{liu2024rmsin} and RefSegRS \cite{yuan2024rrsis} datasets and report the quantitative results in terms of overall IoU, mean IoU, and precision at different thresholds based on various evaluation settings.

\begin{table*}
  \caption{Comparison with existing RIS methods on the RRSIS-D \cite{liu2024rmsin} dataset in terms of oIoU, mIoU, and Pr@$k$.
  The second and third columns report visual encoders and textual encoders, respectively.
  The short names for different encoders (\emph{e.g.,}~ResNet \cite{he2016deepresnet}, Swin Transformer\cite{liu2021swin}) are defined as: $R-101$ and $Swin-B$.
  % $Vis Enc.-Visual\ Encoder, Tex Enc.-Textual\ Encoder$
  The best results are \textbf{bold}.
  % The best results are \textbf{bold} while the second best are \underline{underlined}.
  Note that the performance results of previous methods are taken directly from either RMSIN \cite{liu2024rmsin} or their original works.
  % when available; otherwise, we re-implement the corresponding models to obtain the results (indicated with $\dagger$).
  }
  \centering
  \setlength\tabcolsep{5pt}
  \begin{tabular}{p{2.38cm}cccccccccccc}
  \toprule
  \multirow{2}*{Method} & \multicolumn{1}{c}{Visual} & \multicolumn{1}{c}{Textual} & \multicolumn{2}{c}{Pr@0.5} & \multicolumn{2}{c}{Pr@0.7} & \multicolumn{2}{c}{Pr@0.9} & \multicolumn{2}{c}{oIoU} & \multicolumn{2}{c}{mIoU}\\
 & Encoder & Encoder & Val & Test & Val &  Test & Val &  Test & Val & Test & Val & Test\\
 \midrule
  RRN\cite{li2018rrn} \tiny{\textit{cvpr18}}   & R-101 & LSTM & 51.09 & 51.07 & 33.04 & 32.77 & 6.14 & 6.37 & 66.53 & 66.43 & 46.06 & 45.64\\
  CMSA\cite{ye2019csma} \tiny{\textit{cvpr19}}  & R-101 & None & 55.68 & 55.32 & 38.27 & 37.43 & 9.02 & 8.15 & 69.68 & 69.39 & 48.85 & 48.54\\
  CMPC\cite{huang2020cmpc} \tiny{\textit{cvpr20}} & R-101 & LSTM & 57.93 & 55.83 & 38.50 & 36.94 & 9.31 & 9.19 & 70.15 & 69.22 & 50.41 & 49.24\\
  BRINet \cite{hu2020brinet} \tiny{\textit{cvpr20}} & R-101 & LSTM & 58.79 & 56.90 & 39.65 & 39.12  & 9.19 & 8.73 & 70.73 & 69.88 & 51.14 & 49.65\\
  LSCM\cite{hui2020lscm} \tiny{\textit{eccv20}}  & R-101 & LSTM & 57.12 & 56.02 & 37.87 & 37.70 & 7.93 & 8.27 & 69.28 & 69.05 & 50.36 & 49.92\\
  CMPC+ \cite{liu2021cmpc+} \tiny{\textit{tpami21}} & R-101 & LSTM & 59.19 & 57.65 & 49.36 & 36.97 & 8.16 & 7.78 & 70.14 & 68.64 & 51.41 & 50.24\\
  CRIS \cite{wang2022cris} \tiny{\textit{cvpr22}} & R-101 & CLIP & 56.44 & 54.84 & 39.77 & 38.06 & 11.84 & 11.52 & 70.98 & 70.46 & 50.75 & 49.69\\
  ETRIS \cite{xu2023bridgingclip} \tiny{\textit{iccv23}} & R-101 & CLIP & 62.10 & 61.07 & 43.12 & 40.94 & 12.90 & 11.43 & 72.75 & 71.06 & 55.21 & 54.21\\
  LAVT \cite{yang2022lavt} \tiny{\textit{cvpr22}} & Swin-B & BERT & 69.54 & 69.52 & 53.16 & 53.29 & 24.25 & 24.94 & 77.59 & 77.19 & 61.46 & 61.04\\
  CroVLT \cite{cho2023crossvlt} \tiny{\textit{tmm23}} & Swin-B & BERT & 67.07 & 66.42 & 50.80 & 49.76 & 23.51 & 23.30 & 76.25 & 75.48 & 59.78 & 58.48\\
  CARIS \cite{liu2023caris} \tiny{\textit{mm23}} & Swin-B & BERT & 71.61 & 71.50 & 54.14 & 52.92 & 23.79 & 23.90 & 77.48 & 77.17 & 62.88 & 62.12\\
  LGCE \cite{yuan2024rrsis} \tiny{\textit{tgrs24}} & Swin-B & BERT & 68.10 & 67.65 & 
  52.24 & 51.45 & 23.85 & 23.33 & 76.68 & 76.34 & 60.16 & 59.37\\
  RMSIN \cite{liu2024rmsin} \tiny{\textit{cvpr24}}  & Swin-B & BERT & 74.66 & 74.26 & 57.41 & 55.93 & 24.43 & 24.53 & 78.27 & 77.79 & 65.10 & 64.20\\
  % \hline
  % RMSIN-re40 \cite{liu2024rmsin}  & Swin-B & BERT & 74.48 & 74.06 & 55.57 & 55.39 & 22.59 & 22.61 & 77.63 & 77.01 & 64.81 & 64.03\\
  % RMSIN-re60 \cite{liu2024rmsin}  & Swin-B & BERT & 74.08 & 73.63 & 56.09 & 55.59 & 25.06 & 23.90 & 77.83 & 77.41 & 64.75 & 63.83\\
  % DANet \cite{pan2024mmrrsis} \tiny{\textit{MM24}}  & Swin-B & BERT & 73.69 & - & 57.92 & - & \textbf{27.05} & - & 79.85 & - & 66.07 & -\\
  % \hline
  CroBIM \cite{dong2024crossarxiv} \tiny{\textit{arXiv24}}  & Swin-B & BERT & 74.20 & 75.00 & 54.08 & 54.31 & 22.30 & 21.78 & 76.24 & 76.37 & 63.99 & 64.24\\
  FIANet \cite{lei2024exploringarxiv} \tiny{\textit{arXiv24}} & Swin-B & BERT & - & 74.46 & - & 56.31 & - & 24.13 & - & 76.91 & - & 64.01\\
  SBANet (ours) \tiny{\textit{2024}} & Swin-B & BERT  & \textbf{76.84} & \textbf{75.91} & \textbf{58.86} & \textbf{57.05} & \textbf{26.70} & \textbf{25.38} & \textbf{80.02} & \textbf{79.22} & \textbf{66.71} & \textbf{65.52}\\
  \bottomrule
  \end{tabular}
  \label{tab:comparison with SOTA}
  \vspace{-3mm}
\end{table*}

\subsubsection{Results on the RRSIS-D Dataset}
\label{sec: results on rrsis}
We compared the proposed SBANet with various RIS methods \cite{liu2024rmsin, huang2020cmpc, liu2021cmpc+, yang2022lavt, yuan2024rrsis, lei2024exploringarxiv, liu2023caris, ye2019csma, dong2024crossarxiv, li2018rrn, hu2020brinet, hui2020lscm, wang2022cris, xu2023bridgingclip, cho2023crossvlt} on the RRSIS-D dataset with the adopted evaluation metrics.
The results are reported in Table~\ref{tab:comparison with SOTA}.
We also report visual and textual encoders for each model in this table to indicate the importance of feature extraction in referring segmentation.
From this table, we can observe that the methods with combined Swin Transformer \cite{liu2021swin} and BERT \cite{devlin2018bert} encoders normally outperformed the methods with LSTM \cite{hochreiter1997lstm} or CLIP \cite{radford2021learningclip} models, which achieved at least 5\% improvement over oIoU and mIoU.
Complex semantics and contextual information require high-level understanding of linguistic features, where LSTM, as a simple sequential unit, fails to capture the necessary clues for such a pixel-wise prediction task.
The recent RIS methods mostly adopt the powerful Swin Transformer and BERT as their multi-modal encoders.
Among these models, our proposed SBANet achieved superior performance in terms of oIoU and mIoU.
Specifically, SBANet improved these numbers by 1.43\% and 1.28\% in the testing split, respectively, compared to the previous best-performing methods \cite{liu2024rmsin, dong2024crossarxiv}.
Furthermore, SBANet exhibited more robust results when setting progressively higher thresholds during the evaluation stage.
For example, it improved the results for Pr@0.9 (requiring high-quality segmentation performance) in the validation and testing subsets by 2.27\% and 0.85\%, respectively, compared to RMSIN \cite{liu2024rmsin}.
These results demonstrate the superiority of our proposed scale-wise bidirectional alignment modules and text-conditioned aggregator, which capture global context and local details with updated visual and linguistic features for RRSIS in complex aerial scenarios.

\subsubsection{Results on the RefSegRS Dataset}
\label{sec: results on refsegrs}
We also conducted experiments on the other mainstream benchmark RefSegRS \cite{yuan2024rrsis} and compared the results with previous RIS methods \cite{liu2024rmsin, huang2020cmpc, liu2021cmpc+, yang2022lavt, yuan2024rrsis, liu2023caris, ye2019csma, pan2024mmrrsis, dong2024crossarxiv, li2018rrn, hu2020brinet, cho2023crossvlt}.
As most of these methods only report the results on the testing split, we follow them and show the quantitative comparison in Table~\ref{tab:comparison with SOTA on refsegrs}.
From this table, we can see the similar performance differences in Table~\ref{tab:comparison with SOTA} when comparing different encoders (\emph{e.g.,}~Swin Transformer plus BERT achieve 72\% oIoU at least).
Compared with the previous best-performing methods \cite{liu2024rmsin, pan2024mmrrsis}, our SBANet achieved higher performance with 0.41\% and 1.43\% improvements on Pr@0.5 and PR@0.7, respectively. 
Moreover, it outperformed the second-best model DANet \cite{pan2024mmrrsis} by 0.33\% and 0.59\% for oIoU and mIoU, respectively.
We observe that the improvements were less pronounced compared to the RRSIS-D dataset.
On the one hand, the benefit of feature representation from the proposed bi-directional alignment module is constrained by the quantity and diversity of the RefSegRS dataset.
On the other hand, the referent object in each example from the RefSegRS dataset includes more than one object, which conflicts with the focus designed to be conditioned on linguistic features with specific unique constraints.
For instance, there are some expressions (also shown in Fig.~\ref{fig: refsegrs results}) from RefSegRS such as \textit{vehicle in the parking area} and \textit{van driving on the road}, indicating multiple ground targets from disconnected image regions.
In summary, our proposed SBANet outperformed the previous RIS methods across various evaluation metrics on RefSegRS and can obtain further improvements with additional techniques (\emph{e.g.,}~object-level alignment \cite{liu2023gres}) for multiple referent objects within expressions.

\begin{table}
  \caption{Comparison with existing RIS methods on the RefSegRS \cite{yuan2024rrsis} dataset in terms of oIoU, mIoU, and Pr@$k$.
  The best results are \textbf{bold}.
  Note that the performance results of previous methods are taken directly from either LGCE \cite{yuan2024rrsis} or their original works, when available; otherwise, we re-implement the corresponding models to obtain the results (indicated with $\dagger$).
  }
  \centering
  \setlength\tabcolsep{5pt}
  \begin{tabular}{p{2.1cm}ccccc}
  \toprule
  Method & Pr@0.5 & Pr@0.7 & Pr@0.9 & oIoU & mIoU\\
 % & Val & Test & Val &  Test & Val &  Test & Val & Test & Val & Test\\
 \midrule
  RRN \cite{li2018rrn} & 31.21 & 15.30 & 1.10 & 66.12 & 43.34\\
  CMSA \cite{ye2019csma}  & 28.07 & 12.71 & 0.83 & 64.53 & 41.47 \\
  CMPC \cite{huang2020cmpc} & 26.57 & 11.26 & 0.88 & 61.25 & 33.57 \\
  BRINet \cite{hu2020brinet} & 22.56 & 9.85 & 0.50 & 60.16  & 32.87\\
  % LSCM\cite{hui2020lscm} \tiny{\textit{ECCV20}}  & R-101 & LSTM & 57.12 & 56.02 & 37.87 & 37.70 & 7.93 & 8.27 & 69.28 & 69.05 & 50.36 & 49.92\\
  CMPC+ \cite{liu2021cmpc+} & 51.27 & 29.54 & 3.27 & 68.23 & 54.21 \\
  LAVT \cite{yang2022lavt} & 71.44 & 32.14 & 4.51 & 76.46 & 57.74\\
  CroVLT \cite{cho2023crossvlt} $\dagger$  & 70.58 & 31.29 & 4.16 & 74.89 & 57.02\\
  CARIS \cite{liu2023caris} $\dagger$ & 71.82 & 33.67 & 5.03 & 76.62 & 58.30\\
  LGCE \cite{yuan2024rrsis} & 73.75 & 39.46 & 
  5.45 & 76.81 & 59.96\\
  RMSIN \cite{liu2024rmsin} $\dagger$ & 72.26 & 39.37 & 5.38 & 76.29 & 59.63 \\
  DANet \cite{pan2024mmrrsis} & 76.61 & 42.72 & 8.04 & 79.53 & 62.14 \\
  CroBIM \cite{dong2024crossarxiv} & 64.83 & 17.28 & 2.20 & 72.30 & 52.69 \\
  % FIANet \cite{lei2024exploringarxiv}$\dagger$ & 84.09 & 61.86 & 7.10 & 78.32 & 68.67 \\
  
  SBANet (ours) & \textbf{77.02} & \textbf{44.15} & \textbf{8.97} & \textbf{79.86} & \textbf{62.73}\\
  \bottomrule
  \end{tabular}
  \label{tab:comparison with SOTA on refsegrs}
    \vspace{-2mm}
\end{table}

\subsection{Ablation Studies}
\label{sec: ablation}
To demonstrate the effectiveness of the individual modules proposed in SBANet, we conducted several groups of ablation studies and report the corresponding results in this section.
We followed the settings discussed in Sec.~\ref{sec: details}, except for the variables adjusted for each set of ablation studies.

\subsubsection{Effectiveness of Proposed Modules}
\label{abl: modules}
We quantitatively evaluated the impact of each module proposed in the SBANet on the RRSIS-D \cite{liu2024rmsin} dataset, including the dynamic feature selection block, bidirectional alignment module, and text-conditioned channel and spatial aggregator.
The results are summarized in Table~\ref{tab:ablation of modules}.
We took LAVT \cite{yang2022lavt} as the baseline for a progressive comparison with introduced modules.
To assess the impact of capturing global context and local details at each hierarchical stage for the visual feature refinement, we first added the dynamic feature selection block to the baseline.
The results shown in the second row demonstrate the effectiveness of the block for RRSIS, especially when only requiring a relatively low-quality segmentation performance (\emph{e.g.,}~3.71\% improvement for Pr@0.5).
We subsequently replaced the original vision-language alignment with our proposed bidirectional alignment module to improve the cross-modal interaction.
As shown in the third row, the proposed BAM significantly improved the segmentation performance for both low- and high-quality requirements, thereby boosting the numbers from overall and average aspects.
To assess the proposed aggregator, we only added the module to the baseline.
The corresponding results in the fourth row indicate the crucial role of sufficient information exchange among different scales.
Moreover, when combining all the components together, we obtained the best overall IoU 79.22\% and mean IoU 65.52\%, which demonstrates that these modules can complement each other for RRSIS.
These performance gains evidently highlight the effectiveness of each proposed module in enhancing referring segmentation performance.

\begin{table}
  \caption{Ablation study of each module proposed in our SBANet.
  All models were trained and evaluated on the RRSIS-D dataset.
  % Comparison with existing RIS methods on the RefSegRS \cite{yuan2024rrsis} dataset in terms of oIoU, mIoU, and Pr@$k$.
  We abbreviate the dynamic feature selection block, bidirectional alignment module, and text-conditioned channel and spatial aggregator as DFS, BAM, and TCSA, respectively.
  The best results are \textbf{bold}.
  % $\dagger$ denotes the evaluate performance of our re-implement of the corresponding model.
  }
  \centering
  \setlength\tabcolsep{5pt}
  \begin{tabular}{p{2.4cm}ccccc}
  \toprule
  Method & Pr@0.5 & Pr@0.7 & Pr@0.9 & oIoU & mIoU\\
 % & Val & Test & Val &  Test & Val &  Test & Val & Test & Val & Test\\
 \midrule
  LAVT \cite{yang2022lavt} (baseline) & 69.52 & 53.29 & 24.94 & 77.19 & 61.04\\
  LAVT + DFS & 73.23 & 55.50 & 24.30 & 77.47 & 63.32\\
  LAVT + BAM & 75.63 & 55.96 & 25.19 & 78.00 & 64.93\\
  LAVT + TCSA & 74.58 & 56.30 & 24.97 & 77.44 & 64.27\\
  SBANet (ours) &  \textbf{75.91} & \textbf{57.05} & \textbf{25.38} & \textbf{79.22} & \textbf{65.52}\\
  \bottomrule
  \end{tabular}
  \label{tab:ablation of modules}
    % \vspace{-2mm}
\end{table}

\begin{table}
  \caption{Ablation study of the design for the bidirectional alignment module.
  All models were trained and evaluated on the RRSIS-D dataset.
  Pyramid denotes the hierarchical sizes \{128, 256, 512, 1024\} of learnable query tokens in the four encoder stages.
  }
  \centering
  \setlength\tabcolsep{4pt}
  \begin{tabular}{p{3.4cm}ccccc}
  \toprule
  Method & Pr@0.5 & Pr@0.7 & Pr@0.9 & oIoU & mIoU\\
 % & Val & Test & Val &  Test & Val &  Test & Val & Test & Val & Test\\
 \midrule
  PWAM \cite{yang2022lavt} & 69.52 & 53.29 & 24.94 & 77.19 & 61.04\\
  + self-attention \cite{vaswani2017attention} & 70.84 & 53.40 & 22.84 & 76.30 & 61.44\\
  + cross-attention \cite{liu2023caris} & 70.56 & 53.58 & 22.95 & 76.32 & 61.50\\
  % register  & 75.63 & 55.96 & 24.39 & 78.00 & 64.93\\
  + group-attention \cite{li2021groupformer} & 75.12 & 55.41 & 23.82 & 77.55 & 64.47\\
  + learnable-token (w/o pe) & 74.82 & 55.59 & 24.84 & 77.67 & 64.59\\
  BAM (ours) & \textbf{75.63} & \textbf{55.96} & \textbf{25.19} & \textbf{78.00} & \textbf{64.93}\\
  \hline
  BAM-128 &  72.85 & 54.51 & 24.22 & 76.74 & 63.71\\
  BAM-225 (ours) & \textbf{75.63} & \textbf{55.96} & 25.19 & 78.00 & \textbf{64.93}\\ 
  BAM-256 & 75.58 & 55.95 & \textbf{25.21} & \textbf{78.01} & 64.87\\
  BAM-512 & 75.31 & 55.22 & 24.62 & 77.26 & 64.30\\
  BAM-1024 & 74.55 & 55.05 & 24.30 & 76.92 & 64.07\\
  BAM-pyramid & 74.37 & 54.16 & 24.16 & 76.55 & 63.43\\
  \bottomrule
  \end{tabular}
  \label{tab:ablation of bam}
    \vspace{-2mm}
\end{table}

\subsubsection{Influence of Bidirectional Alignment Module}
\label{abl: bam}
To better understand the design of the proposed bidirectional alignment module, we further compared various options with ours for updating visual and linguistic features during cross-modal interaction stage.
The corresponding results are reported in Table~\ref{tab:ablation of bam}.
We adopted PWAM from our baseline LAVT \cite{yang2022lavt} as the default one, which utilizes the language-aware signal to only refine visual features.
To further refine linguistic features, we first added a vanilla self-attention \cite{vaswani2017attention} to softly process the textual focus.
Alternatively, we employed the opposite cross-attention from vision to language inspired by CARIS \cite{liu2023caris}.
However, from the second and third rows in the upper section, both two options only obtained marginal performance improvements over the baseline.
The results indicate that treating each element in visual or linguistic features equally fails to effectively capture the guidance for cross-modal interaction.
We further modified the recent group-attention \cite{li2021groupformer, tang2023contrastivegroup} for updating linguistic features with the visual guidance, which utilized a clustering strategy to group visual features.
Subsequently, we provided two variants (with or without positional embedding) of our BAM with learnable query tokens for sparsely visual representation.
From the last three rows in the upper section, it is evident that applying selective representations for cross-modal interaction improved referring segmentation performance.
Compared to the group attention, our proposed method achieved superior results across all the metrics and avoided the inconvenience and instability brought by their adopted non-differentiable assignment with Gumbel-softmax \cite{jang2016categoricalgumbel, maddison2016concretegumbel}.
Lastly, we report the results based on different numbers of learnable query tokens in the bottom section of Table~\ref{tab:ablation of bam}.
% Too ample or inadequate learnable tokens failed to effectively provide visual features and resulted in poor performance.
Excessive or insufficient learnable tokens failed to effectively capture visual features, leading to poor performance.
A pyramid group for the hierarchical encoder did not enhance the results, either, as shown in the last row.
Thus, we fixed 225 (15 $\times$ 15 from the last stage of the encoder) as our default number of learnable query tokens for each BAM.

\begin{figure*}[htp]
  \centering
  \begin{subfigure}[t]{1\linewidth}
  \subcaption*{\textbf{$Expression$:} The oval large green and red ground track field}
  \end{subfigure}
 
  \begin{subfigure}[t]{0.19\linewidth}
  \includegraphics[width=0.95\linewidth]{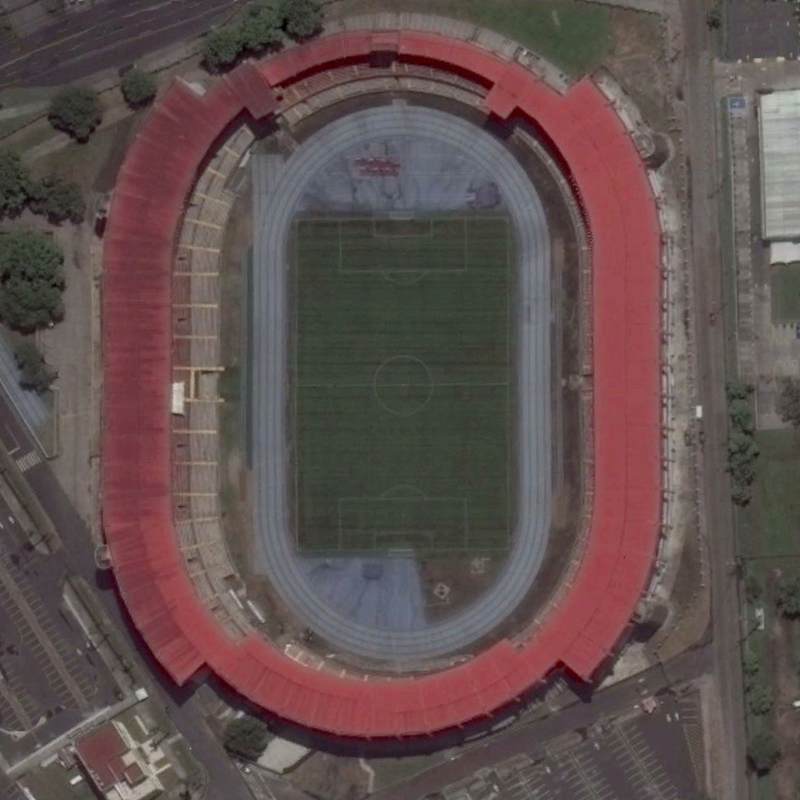}
  \end{subfigure}
  \begin{subfigure}[t]{0.19\linewidth}
  \includegraphics[width=0.95\linewidth]{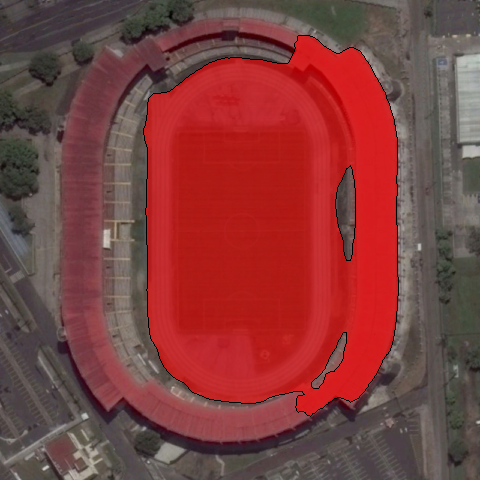}
  \end{subfigure}
  \begin{subfigure}[t]{0.19\linewidth}
  \includegraphics[width=0.95\linewidth]{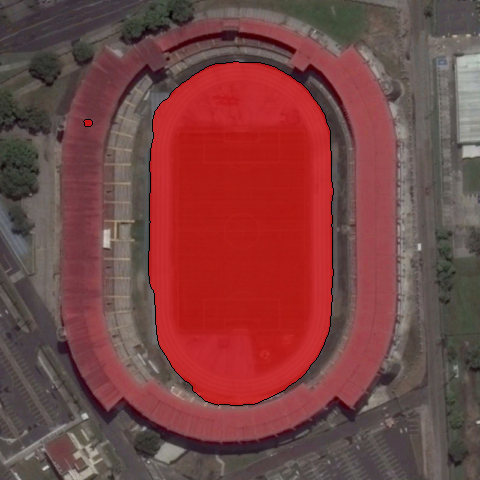}
  \end{subfigure}
  \begin{subfigure}[t]{0.19\linewidth}
  \includegraphics[width=0.95\linewidth]{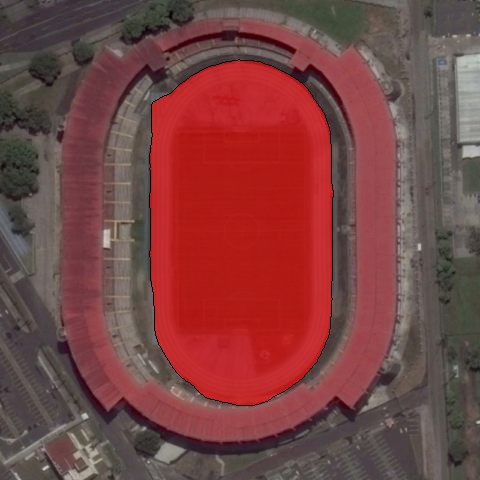}
  \end{subfigure}
  \begin{subfigure}[t]{0.19\linewidth}
  \includegraphics[width=0.95\linewidth]{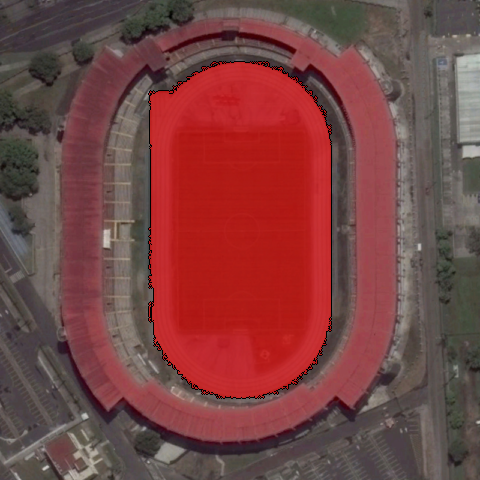}
  \end{subfigure}

  \vspace{-5mm} 

  \begin{subfigure}[t]{1\linewidth}
  \subcaption*{\textbf{$Expression$:} A harbor in the middle}
  \end{subfigure}
  
  \begin{subfigure}[t]{0.19\linewidth}
  \includegraphics[width=0.95\linewidth]{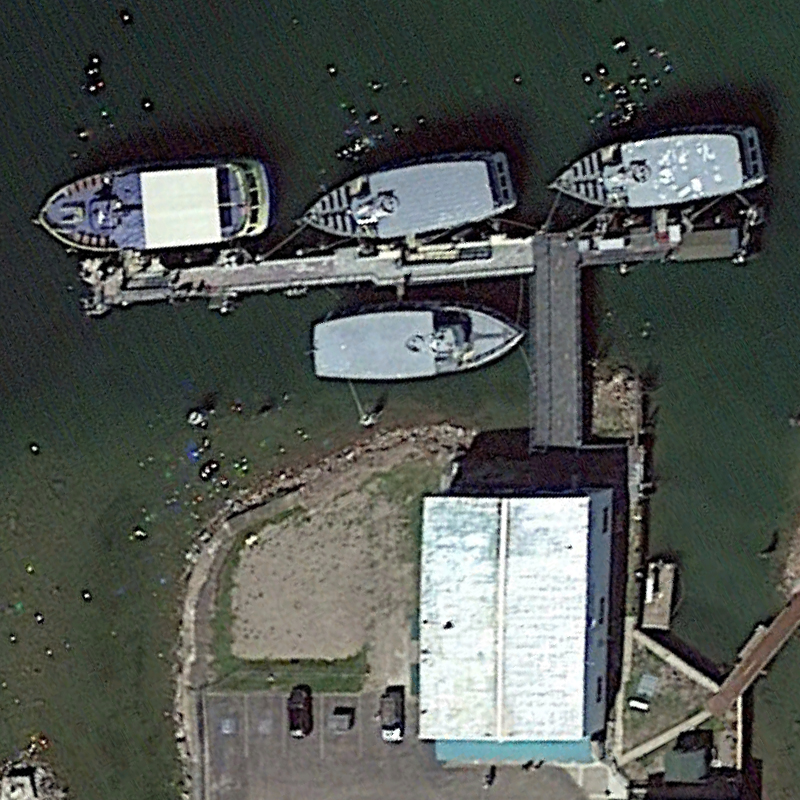}
  \end{subfigure}
  \begin{subfigure}[t]{0.19\linewidth}
  \includegraphics[width=0.95\linewidth]{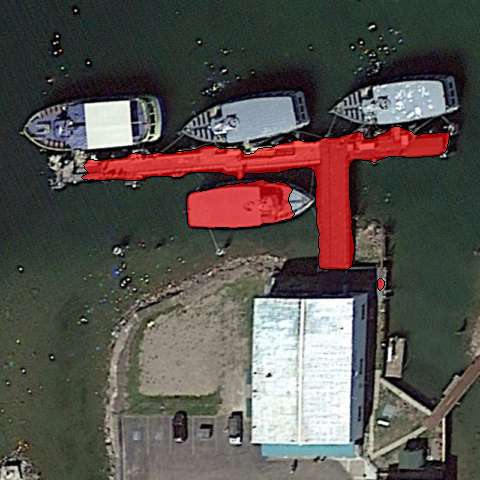}
  \end{subfigure}
  \begin{subfigure}[t]{0.19\linewidth}
  \includegraphics[width=0.95\linewidth]{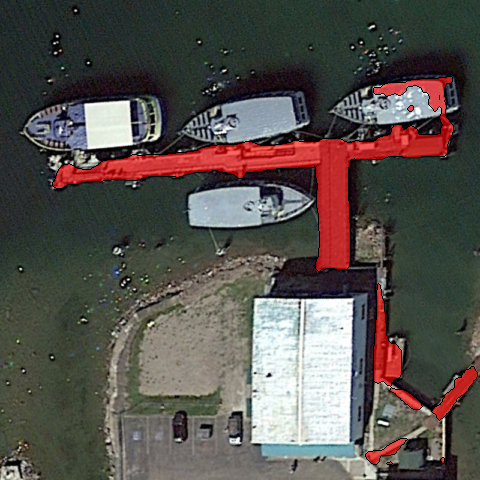}
  \end{subfigure}
  \begin{subfigure}[t]{0.19\linewidth}
  \includegraphics[width=0.95\linewidth]{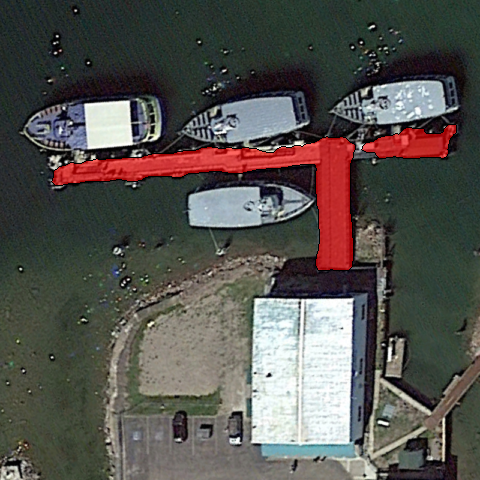}
  \end{subfigure}
  \begin{subfigure}[t]{0.19\linewidth}
  \includegraphics[width=0.95\linewidth]{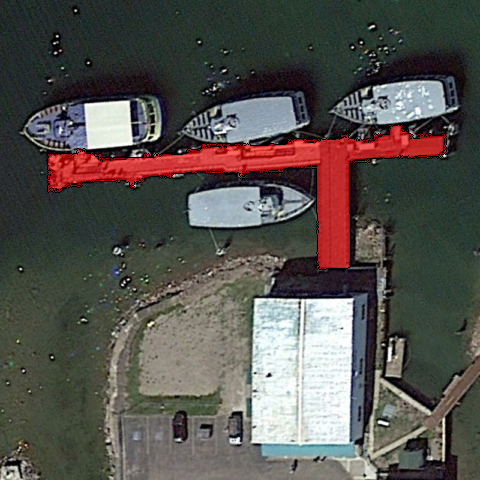}
  \end{subfigure}

  \vspace{-5mm} 
  
  \begin{subfigure}[t]{1\linewidth}
  \subcaption*{\textbf{$Expression$:} The airplane on the left}
  \end{subfigure}
  
  \begin{subfigure}[t]{0.19\linewidth}
  \includegraphics[width=0.95\linewidth]{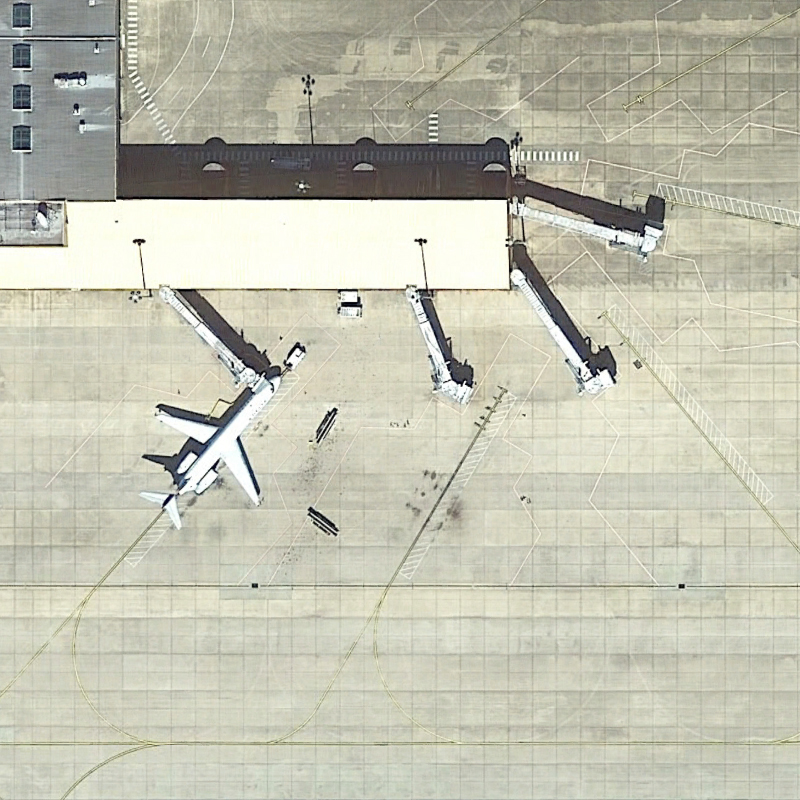}
  \end{subfigure}
  \begin{subfigure}[t]{0.19\linewidth}
  \includegraphics[width=0.95\linewidth]{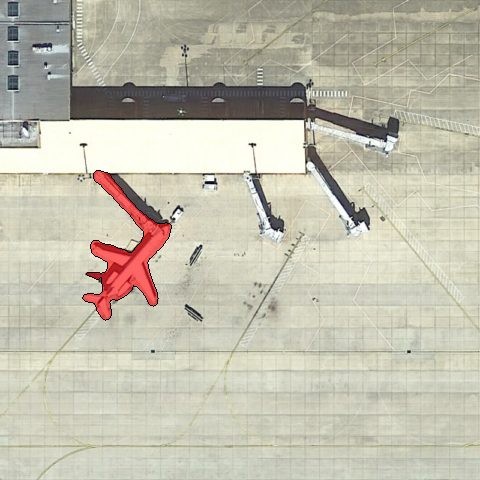}
  \end{subfigure}
  \begin{subfigure}[t]{0.19\linewidth}
  \includegraphics[width=0.95\linewidth]{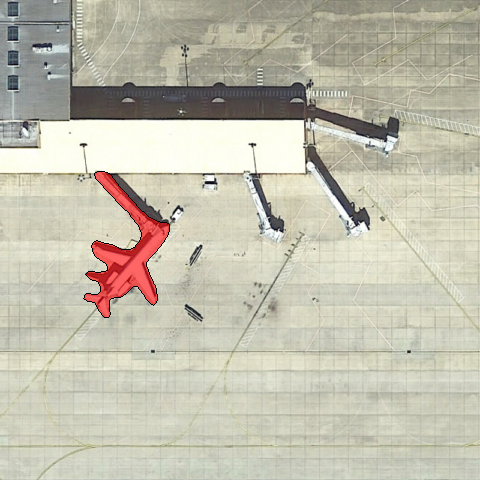}
  \end{subfigure}
  \begin{subfigure}[t]{0.19\linewidth}
  \includegraphics[width=0.95\linewidth]{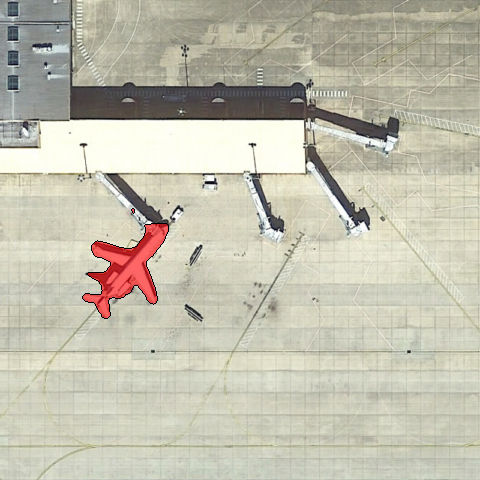}
  \end{subfigure}
  \begin{subfigure}[t]{0.19\linewidth}
  \includegraphics[width=0.95\linewidth]{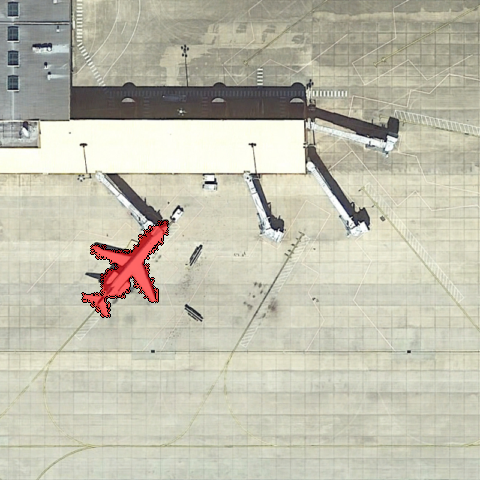}
  \end{subfigure}

  \vspace{-5mm} 
  
  \begin{subfigure}[t]{1\linewidth}
  \subcaption*{\textbf{$Expression$:} A small airport}
  \end{subfigure}
  
  \begin{subfigure}[t]{0.19\linewidth}
  \includegraphics[width=0.95\linewidth]{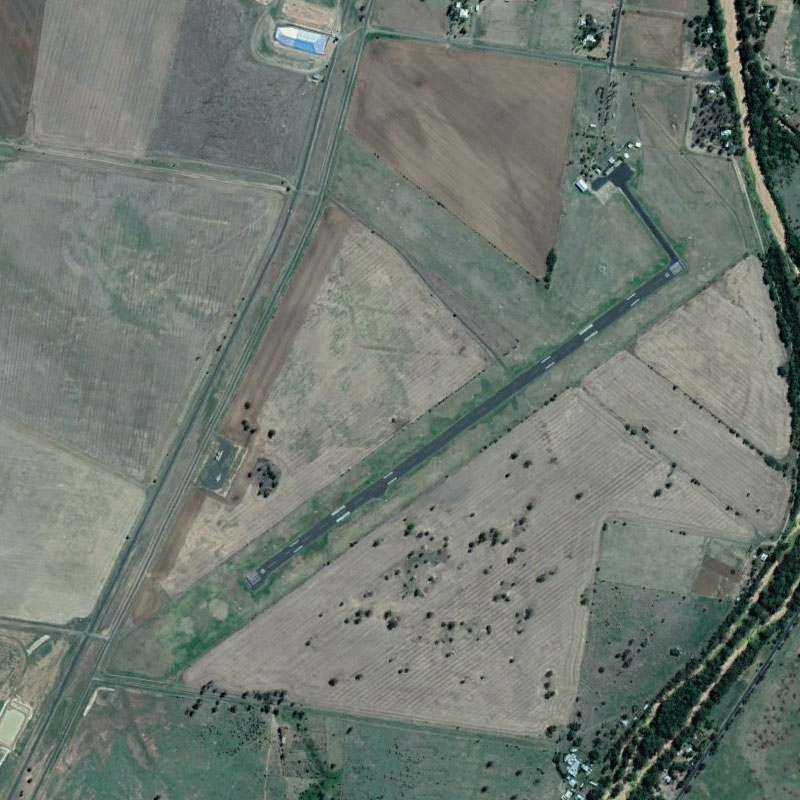}
  \end{subfigure}
  \begin{subfigure}[t]{0.19\linewidth}
  \includegraphics[width=0.95\linewidth]{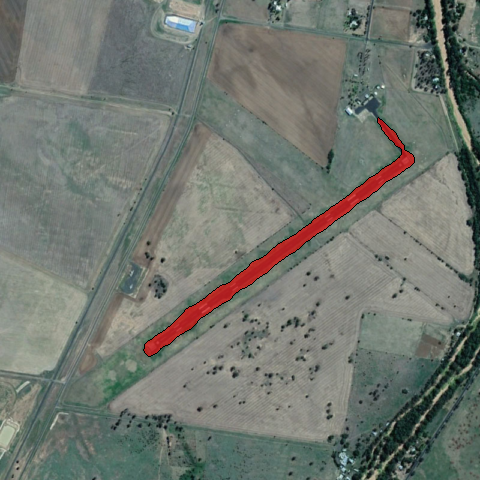}
  \end{subfigure}
  \begin{subfigure}[t]{0.19\linewidth}
  \includegraphics[width=0.95\linewidth]{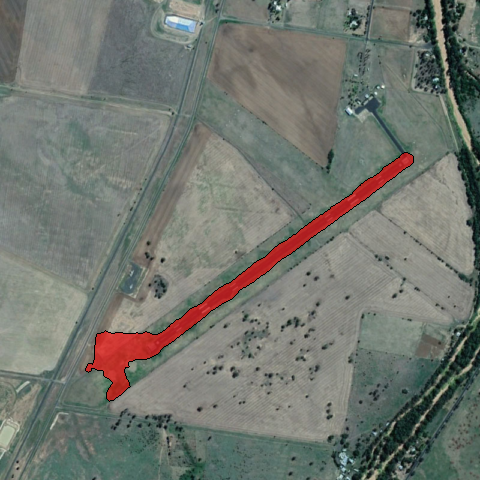}
  \end{subfigure}
  \begin{subfigure}[t]{0.19\linewidth}
  \includegraphics[width=0.95\linewidth]{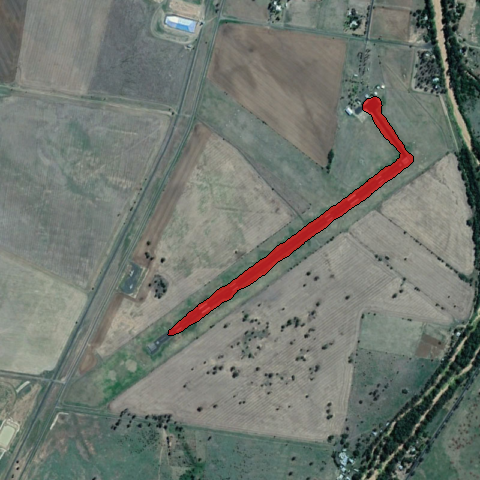}
  \end{subfigure}
  \begin{subfigure}[t]{0.19\linewidth}
  \includegraphics[width=0.95\linewidth]{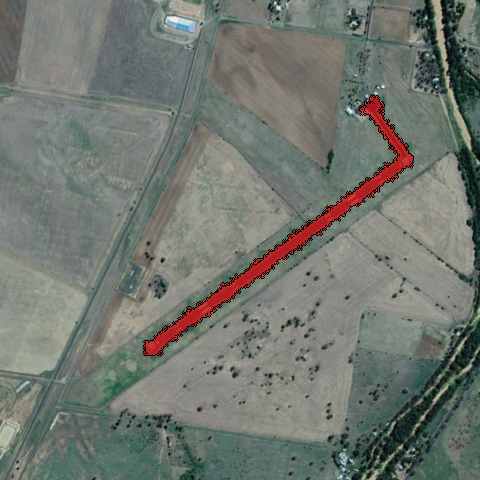}
  \end{subfigure}

  \vspace{-5mm} 

  \begin{subfigure}[t]{1\linewidth}
  \subcaption*{\textbf{$Expression$:} The dam on the top}
  \end{subfigure}
  
  \begin{subfigure}[t]{0.19\linewidth}
  \includegraphics[width=0.95\linewidth]{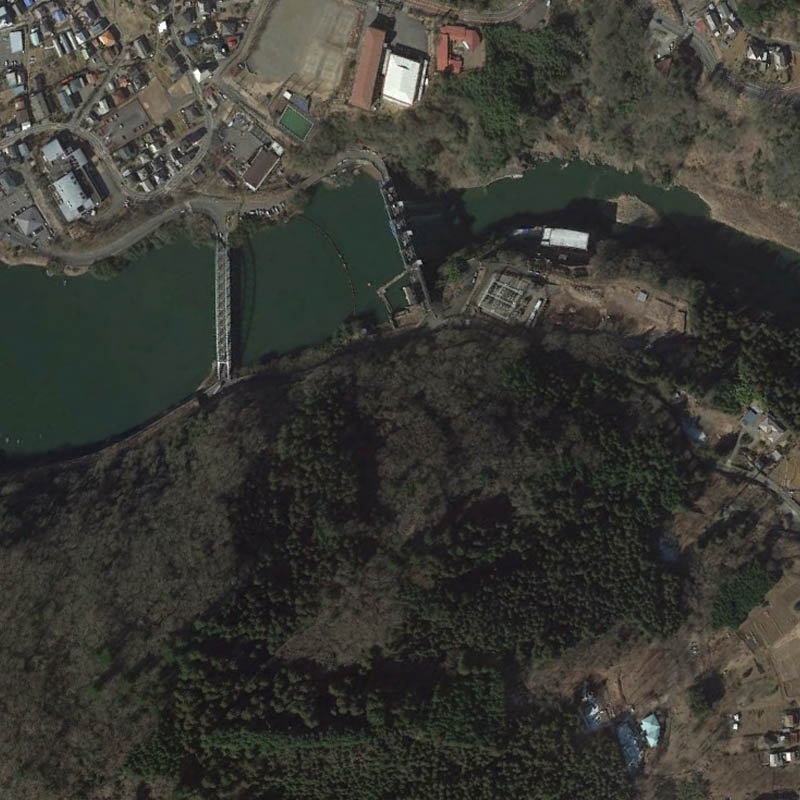}
  \end{subfigure}
  \begin{subfigure}[t]{0.19\linewidth}
  \includegraphics[width=0.95\linewidth]{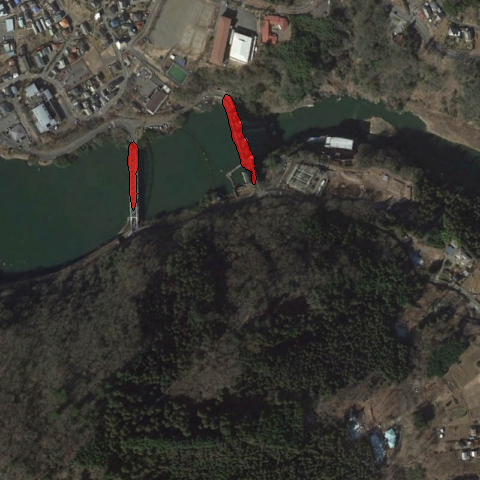}
  \end{subfigure}
  \begin{subfigure}[t]{0.19\linewidth}
  \includegraphics[width=0.95\linewidth]{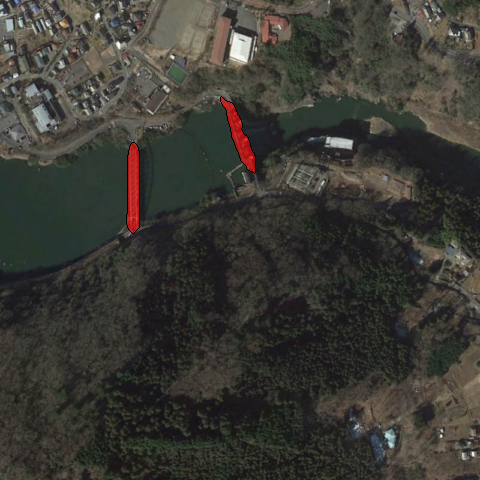}
  \end{subfigure}
  \begin{subfigure}[t]{0.19\linewidth}
  \includegraphics[width=0.95\linewidth]{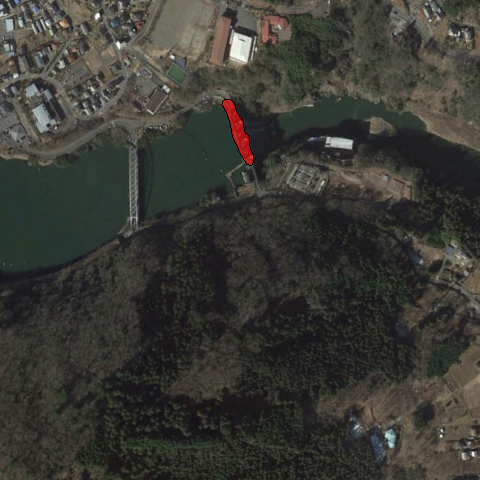}
  \end{subfigure}
  \begin{subfigure}[t]{0.19\linewidth}
  \includegraphics[width=0.95\linewidth]{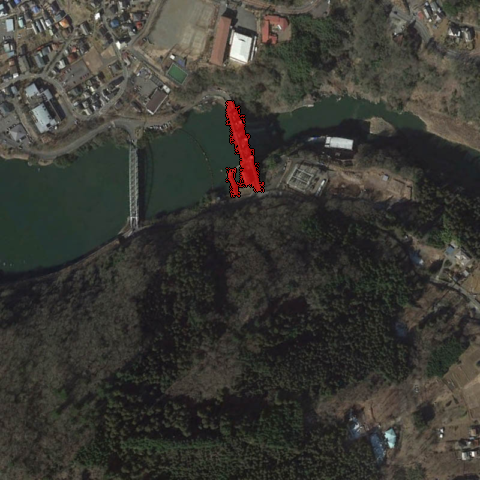}
  \end{subfigure}

  \vspace{-5mm}

  \begin{subfigure}[t]{1\linewidth}
  \subcaption*{\textbf{$Expression$:} A blue vehicle}
  \end{subfigure}
  
  \begin{subfigure}[t]{0.19\linewidth}
  \includegraphics[width=0.95\linewidth]{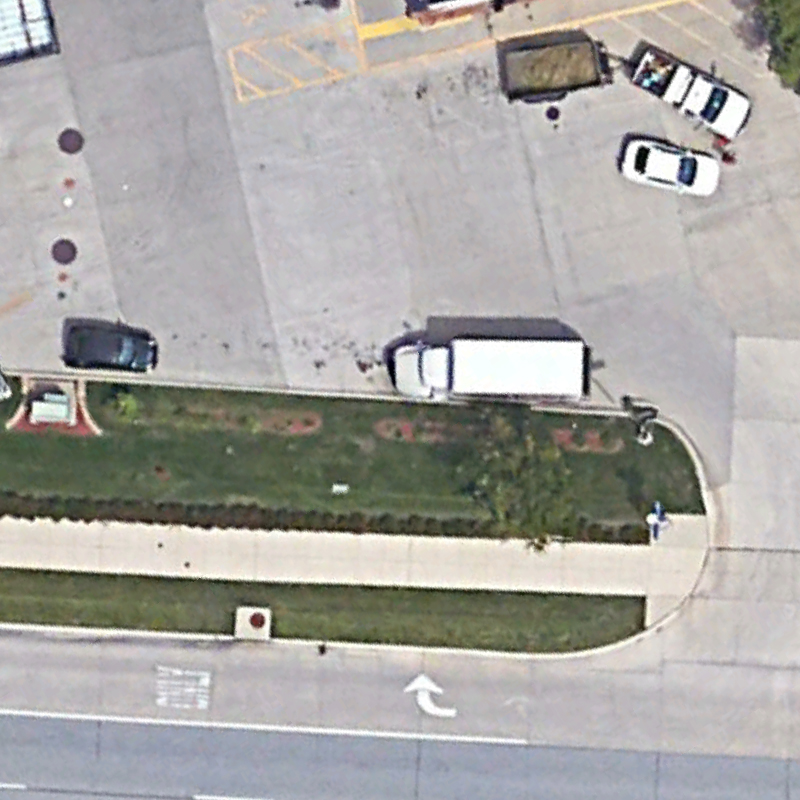}
  \subcaption*{Input Image}
  \end{subfigure}
  \begin{subfigure}[t]{0.19\linewidth}
  \includegraphics[width=0.95\linewidth]{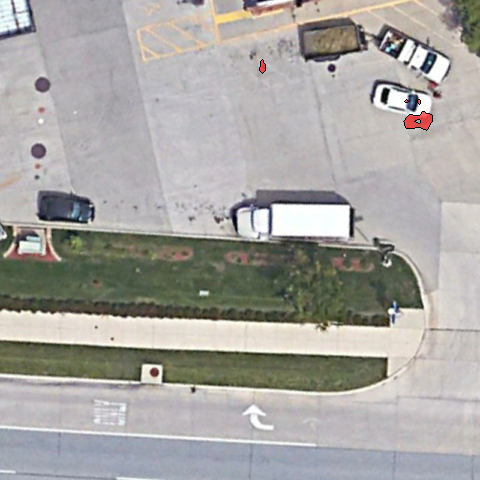}
  \subcaption*{LAVT}
  \end{subfigure}
  \begin{subfigure}[t]{0.19\linewidth}
  \includegraphics[width=0.95\linewidth]{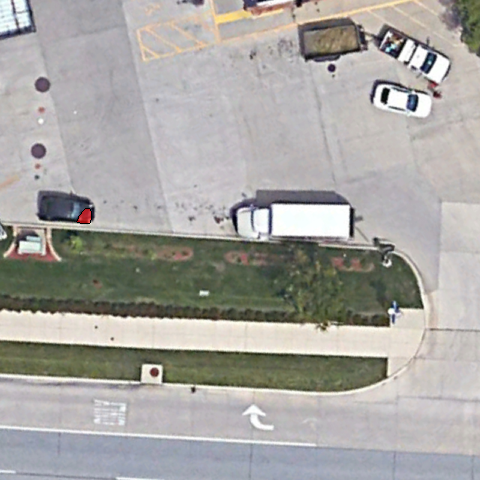}
  \subcaption*{RMSIN}
  \end{subfigure}
  \begin{subfigure}[t]{0.19\linewidth}
  \includegraphics[width=0.95\linewidth]{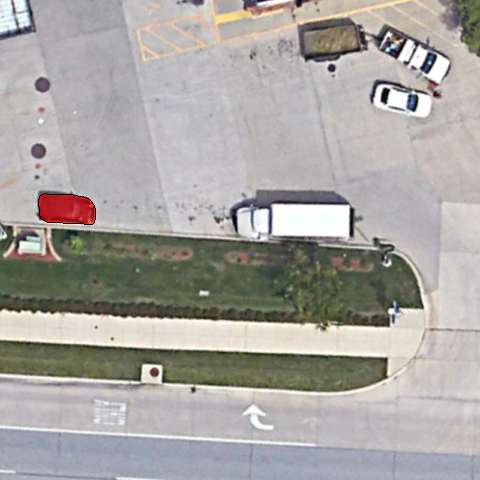}
  \subcaption*{SBANet (ours)}
  \end{subfigure}
  \begin{subfigure}[t]{0.19\linewidth}
  \includegraphics[width=0.95\linewidth]{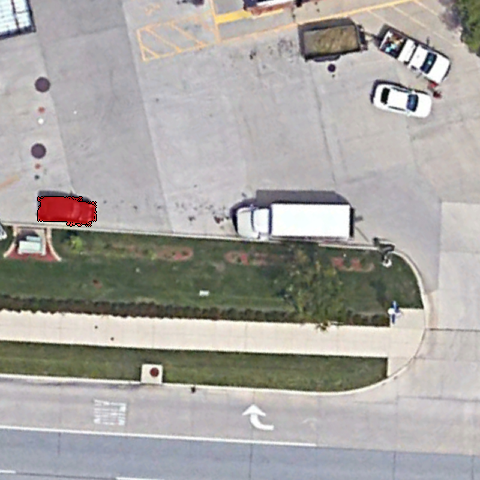}
  \subcaption*{Ground Truth}
  \end{subfigure}
  
  \vspace{-2mm}
  \caption{Qualitative comparison of different methods on the RRSIS-D \cite{liu2024rmsin} dataset (Best view in Zoom).
  }
  \vspace{-5mm}
  \label{fig: rrsis results}
\end{figure*}

\subsubsection{Structure Analysis of Aggregator}
\label{abl: aggregator}
To validate the effectiveness of the proposed text-conditioned channel and spatial aggregator, we conducted ablation experiments on its main components.
The results are summarized in Table~\ref{tab:ablation of aggregator}.
The default method did not adopt any special operations to bridge the encoder and decoder, just like LAVT \cite{yang2022lavt} did.
We first separately added the channel aggregator and spatial aggregator to the default one.
The second and third rows indicate that the model achieved 2.53\% and 1.06\% improvements on mean IoU when only applying one of them.
Subsequently, we combined them together in a sequence and the corresponding results were further improved (2.93\%).
Lastly, we completed the proposed aggregator with the textual guidance from the updated linguistic features.
The method obtained 77.44\% and 64.27\% for overall IoU and mean IoU, respectively, significantly surpassing the previous LAVT model.
Involving textual guidance in the cross-scale information exchange with channel and spatial enhancement helps to capture crucial focuses, resulting in better segmentation performance.

\begin{table}
  \caption{Ablation study of the design for the text-conditioned channel and spatial aggregator.
  All models were trained and evaluated on the RRSIS-D dataset.
  }
  \centering
  \setlength\tabcolsep{4pt}
  \begin{tabular}{p{2.4cm}ccccc}
  \toprule
  Method & Pr@0.5 & Pr@0.7 & Pr@0.9 & oIoU & mIoU\\
 \midrule
  Default & 69.52 & 53.29 & 24.94 & 77.19 & 61.04\\
  % + CIM \cite{liu2024rmsin} & 73.68 & 56.67 & 25.69 & 77.40 & 64.25\\
  + channel & 72.22 & 54.47 & 24.25 & \textbf{77.45} & 63.57\\
  + spatial & 72.65 & 54.04 & 23.18 & 77.16 & 62.10\\
  TCSA (w/o text) & 73.94 & 55.08 & 23.33 & 77.21 & 63.97\\
  TCSA (ours) & \textbf{74.58} & \textbf{56.30} & \textbf{24.97} & 77.44 & \textbf{64.27}\\
  \bottomrule
  \end{tabular}
  \label{tab:ablation of aggregator}
    % \vspace{-2mm}
\end{table}

\begin{table}
  \caption{Ablation study of encoders for input images and expressions.
  All models were trained and evaluated on the RRSIS-D dataset.
  }
  \centering
  \setlength\tabcolsep{4pt}
  \begin{tabular}{ccccccc}
  \toprule
  Visual & Textual & Pr@0.5 & Pr@0.7 & Pr@0.9 & oIoU & mIoU\\
 \midrule
  R-101 & LSTM & 60.06 & 39.10 & 8.04 & 69.71 & 49.88\\
  R-101 & CLIP & 62.11 & 41.17 & 12.00 & 71.63 & 55.29\\
  % R-101 & BERT & 73.23 & 55.50 & 24.30 & 77.47 & 63.32\\
  Swin-B & BERT & \textbf{75.91} & \textbf{57.05} & \textbf{25.38} & \textbf{79.22} & \textbf{65.52}\\
  \bottomrule
  \end{tabular}
  \label{tab:ablation of encoders}
    \vspace{-2mm}
\end{table}

\subsubsection{Backbone Analysis of Encoders}
\label{abl: encoders}
We analyzed different encoders for the feature extraction and evaluated the generalization capability of our proposed modules.
As shown in Table~\ref{tab:ablation of encoders}, three kinds of combination are reported: (a) ResNet-101 \cite{he2016deepresnet} and LSTM \cite{hochreiter1997lstm}; (b) ResNet-101 and CLIP \cite{radford2021learningclip}; (c) Swin Transformer \cite{liu2021swin} and BERT \cite{devlin2018bert}.
We observe the similar performance differences as in Table~\ref{tab:comparison with SOTA}, which (c) outperformed the others for RIS.
Moreover, our designed modules built on (a) and (b) still achieved competitive results compared to previous methods of the corresponding integration in Table~\ref{tab:comparison with SOTA}, which demonstrate the superiority of our designs.

\begin{figure*}[htp]
  \centering
  \begin{subfigure}[t]{1\linewidth}
  \subcaption*{\textbf{$Expression$:} van driving on the road}
  \end{subfigure}
 
  \begin{subfigure}[t]{0.19\linewidth}
  \includegraphics[width=0.95\linewidth]{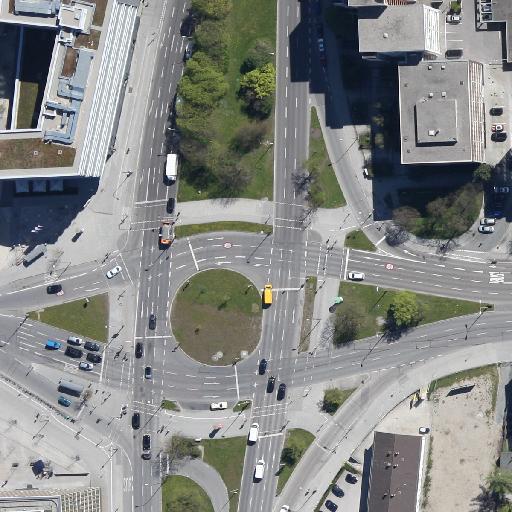}
  \end{subfigure}
  \begin{subfigure}[t]{0.19\linewidth}
  \includegraphics[width=0.95\linewidth]{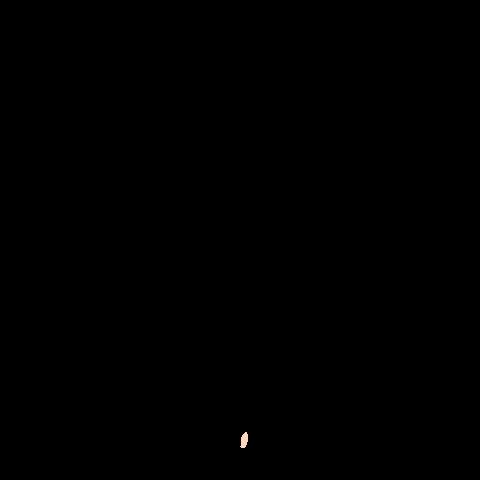}
  \end{subfigure}
  \begin{subfigure}[t]{0.19\linewidth}
  \includegraphics[width=0.95\linewidth]{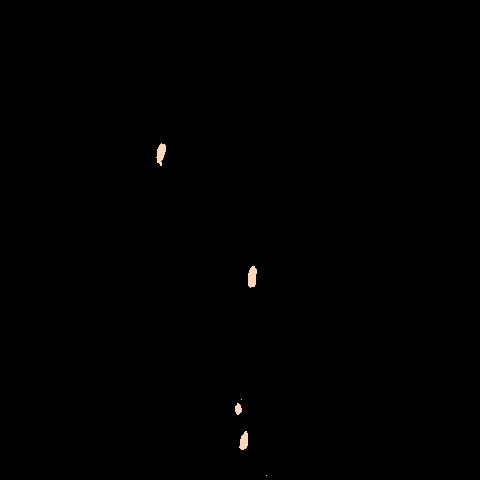}
  \end{subfigure}
  \begin{subfigure}[t]{0.19\linewidth}
  \includegraphics[width=0.95\linewidth]{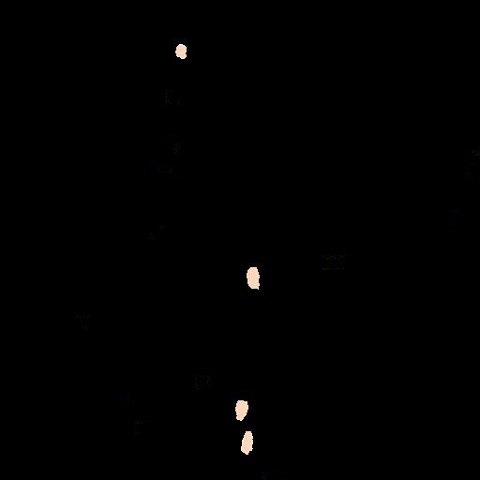}
  \end{subfigure}
  \begin{subfigure}[t]{0.19\linewidth}
  \includegraphics[width=0.95\linewidth]{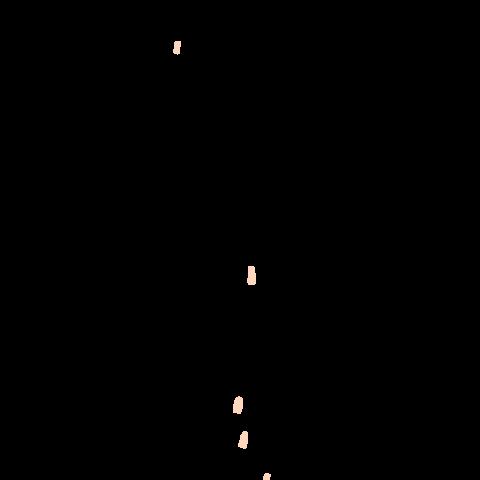}
  \end{subfigure}

  \vspace{-5mm} 
  
  \begin{subfigure}[t]{1\linewidth}
  \subcaption*{\textbf{$Expression$:} paved road}
  \end{subfigure}
  
  \begin{subfigure}[t]{0.19\linewidth}
  \includegraphics[width=0.95\linewidth]{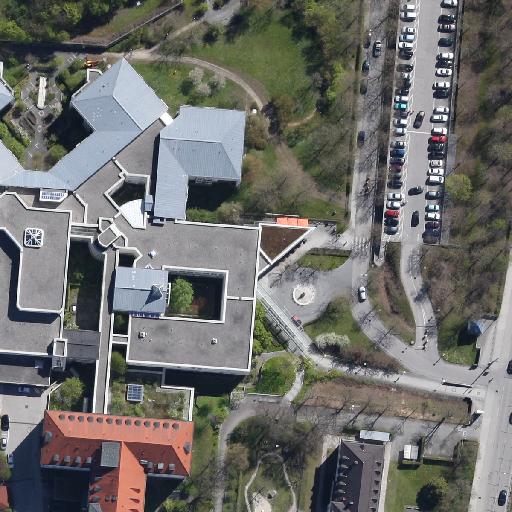}
  \end{subfigure}
  \begin{subfigure}[t]{0.19\linewidth}
  \includegraphics[width=0.95\linewidth]{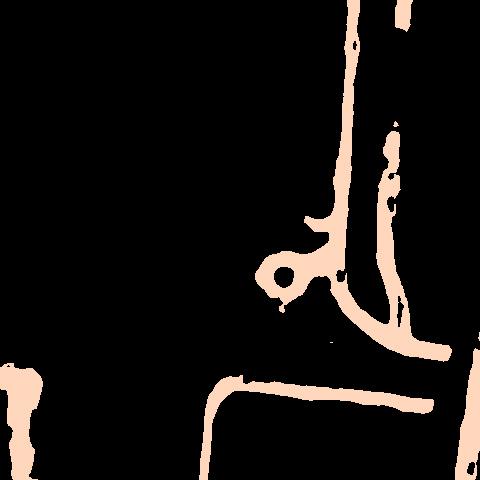}
  \end{subfigure}
  \begin{subfigure}[t]{0.19\linewidth}
  \includegraphics[width=0.95\linewidth]{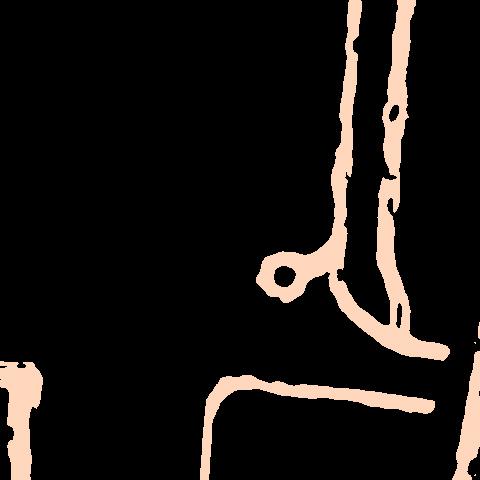}
  \end{subfigure}
  \begin{subfigure}[t]{0.19\linewidth}
  \includegraphics[width=0.95\linewidth]{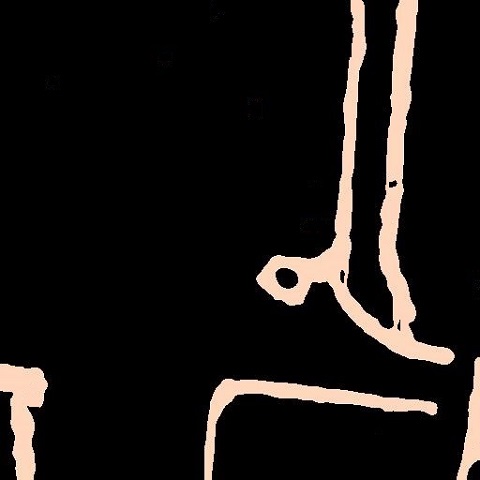}
  \end{subfigure}
  \begin{subfigure}[t]{0.19\linewidth}
  \includegraphics[width=0.95\linewidth]{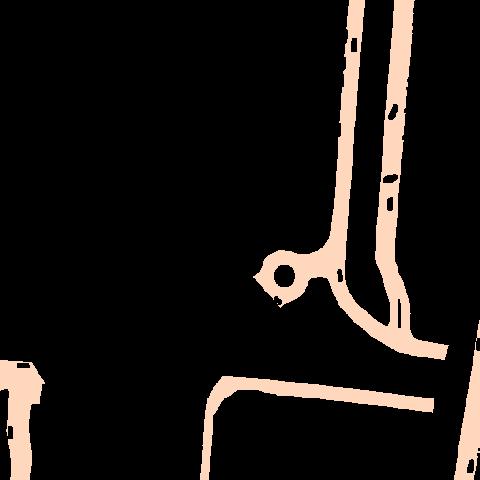}
  \end{subfigure}

  \vspace{-5mm} 

  \begin{subfigure}[t]{1\linewidth}
  \subcaption*{\textbf{$Expression$:} building along the road}
  \end{subfigure}
  
  \begin{subfigure}[t]{0.19\linewidth}
  \includegraphics[width=0.95\linewidth]{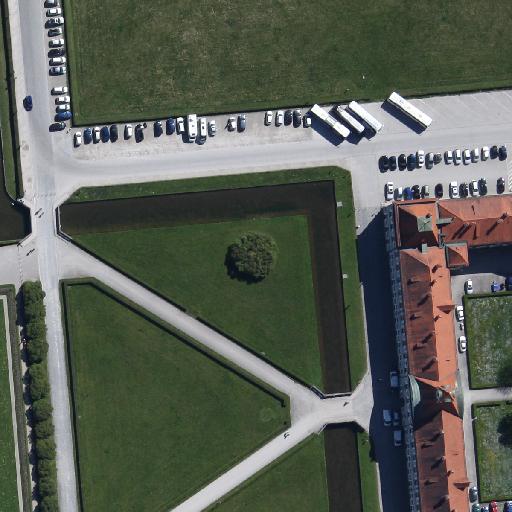}
  \subcaption*{Input Image}
  \end{subfigure}
  \begin{subfigure}[t]{0.19\linewidth}
  \includegraphics[width=0.95\linewidth]{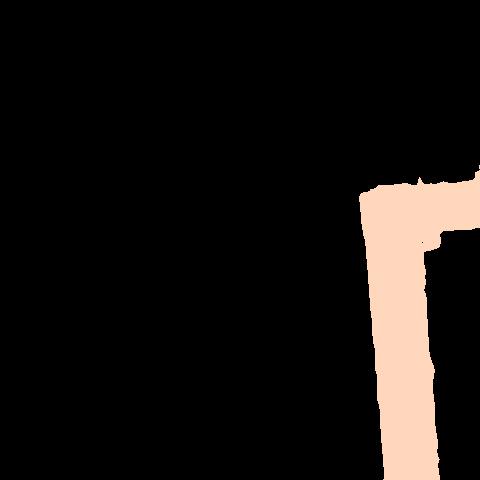}
  \subcaption*{LAVT}
  \end{subfigure}
  \begin{subfigure}[t]{0.19\linewidth}
  \includegraphics[width=0.95\linewidth]{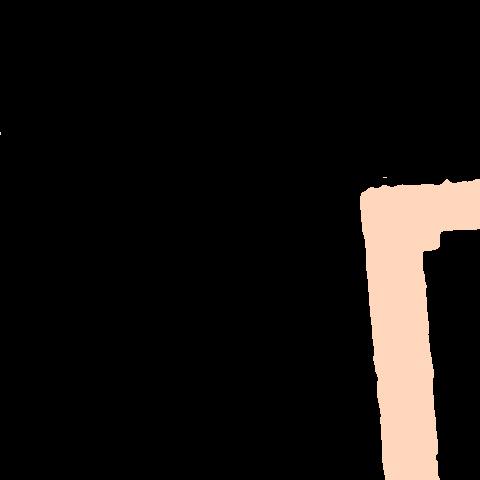}
  \subcaption*{LGCE}
  \end{subfigure}
  \begin{subfigure}[t]{0.19\linewidth}
  \includegraphics[width=0.95\linewidth]{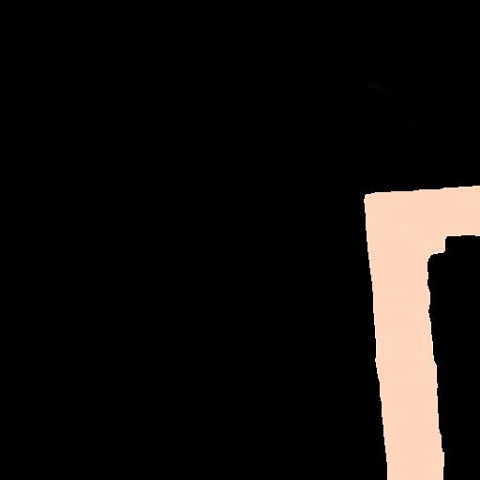}
  \subcaption*{SBANet (ours)}
  \end{subfigure}
  \begin{subfigure}[t]{0.19\linewidth}
  \includegraphics[width=0.95\linewidth]{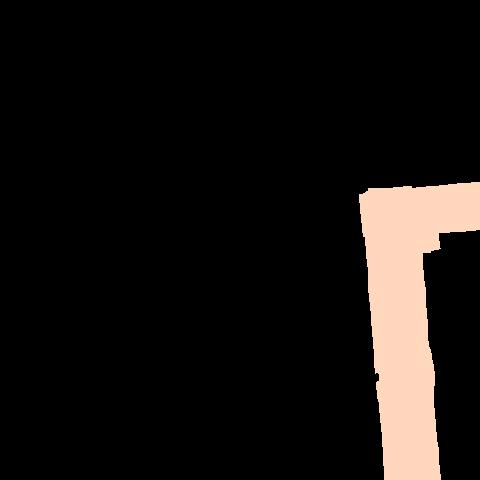}
  \subcaption*{Ground Truth}
  \end{subfigure}
  
  \vspace{-2mm}
  \caption{Qualitative comparison of different methods on the RefSegRS \cite{yuan2024rrsis} dataset (Best view in Zoom).
  }
  \vspace{-5mm}
  \label{fig: refsegrs results}
\end{figure*}

\begin{figure*}[htp]
  \centering
  \begin{subfigure}[t]{1\linewidth}
  \subcaption*{\textbf{$Expression$:} The tennis court on the right}
  \end{subfigure}
 
  \begin{subfigure}[t]{0.19\linewidth}
  \includegraphics[width=0.95\linewidth]{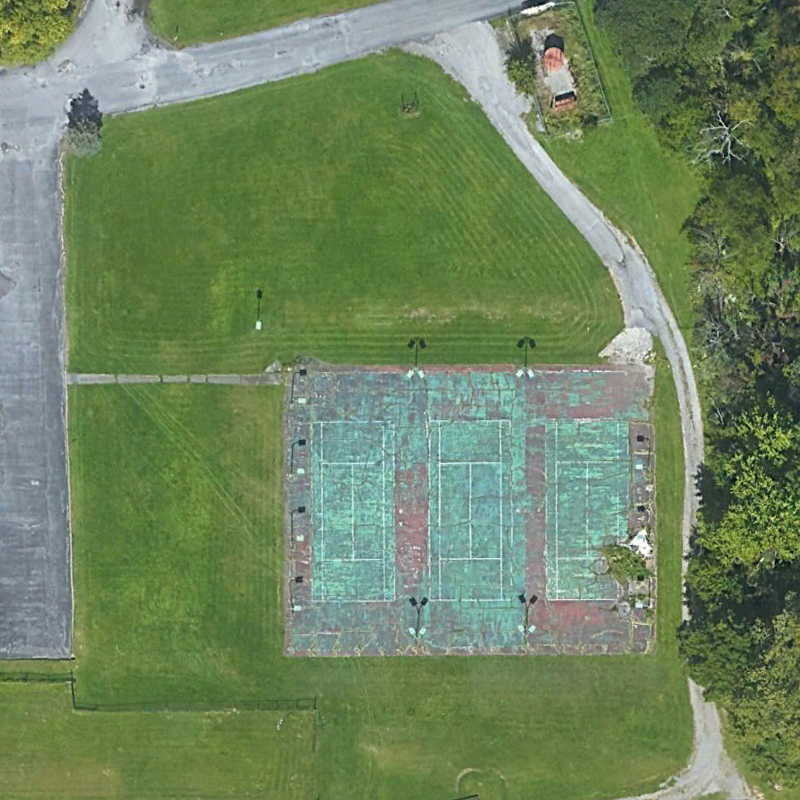}
  \end{subfigure}
  \begin{subfigure}[t]{0.19\linewidth}
  \includegraphics[width=0.95\linewidth]{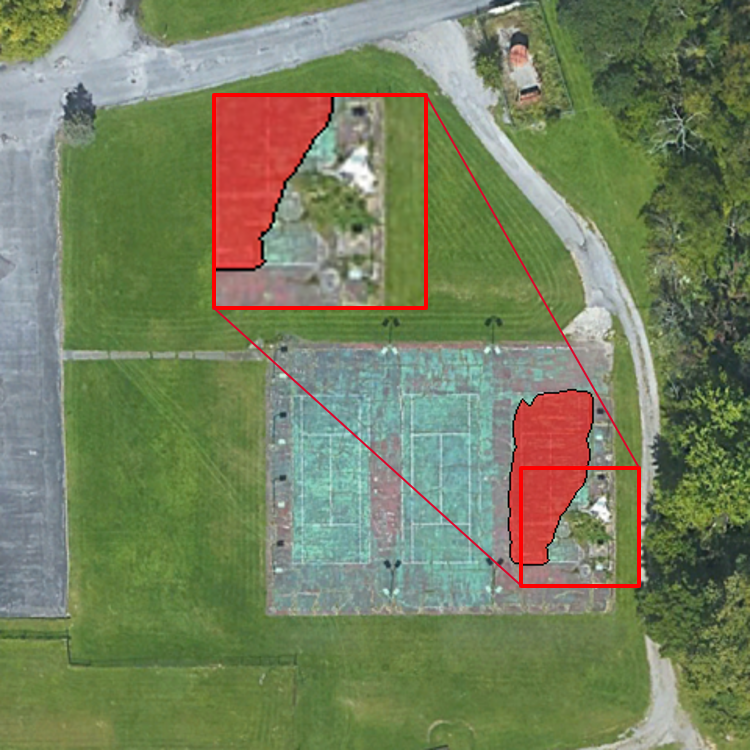}
  \end{subfigure}
  \begin{subfigure}[t]{0.19\linewidth}
  \includegraphics[width=0.95\linewidth]{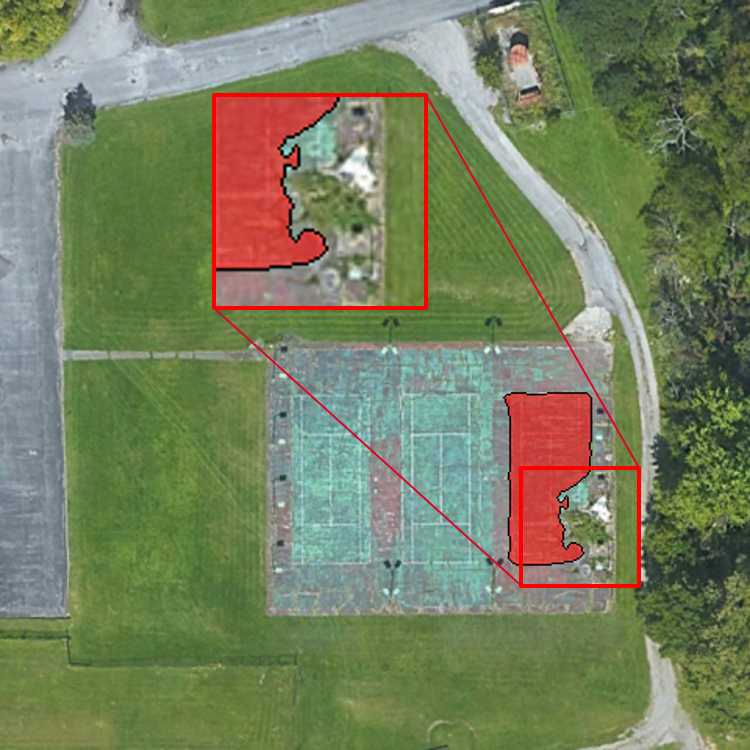}
  \end{subfigure}
  \begin{subfigure}[t]{0.19\linewidth}
  \includegraphics[width=0.95\linewidth]{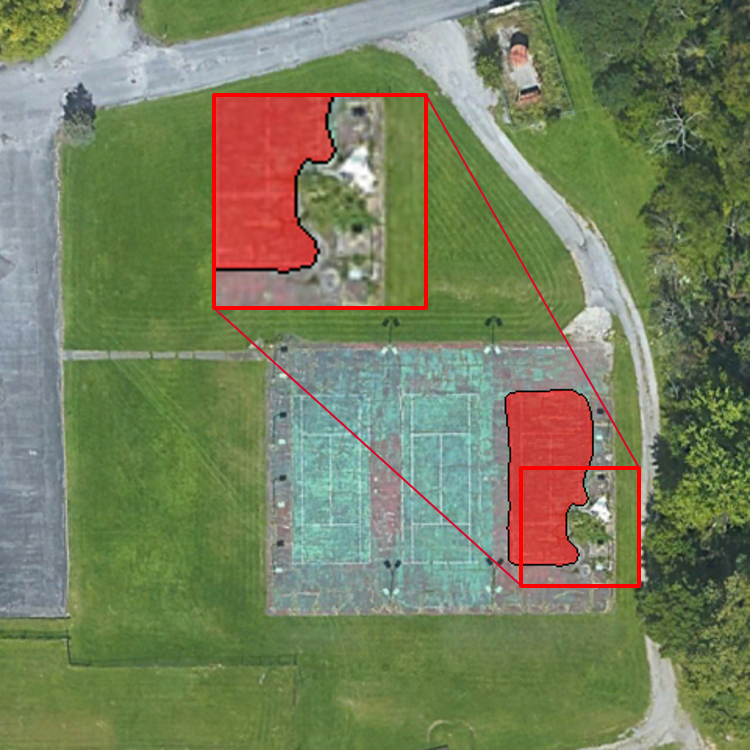}
  \end{subfigure}
  \begin{subfigure}[t]{0.19\linewidth}
  \includegraphics[width=0.95\linewidth]{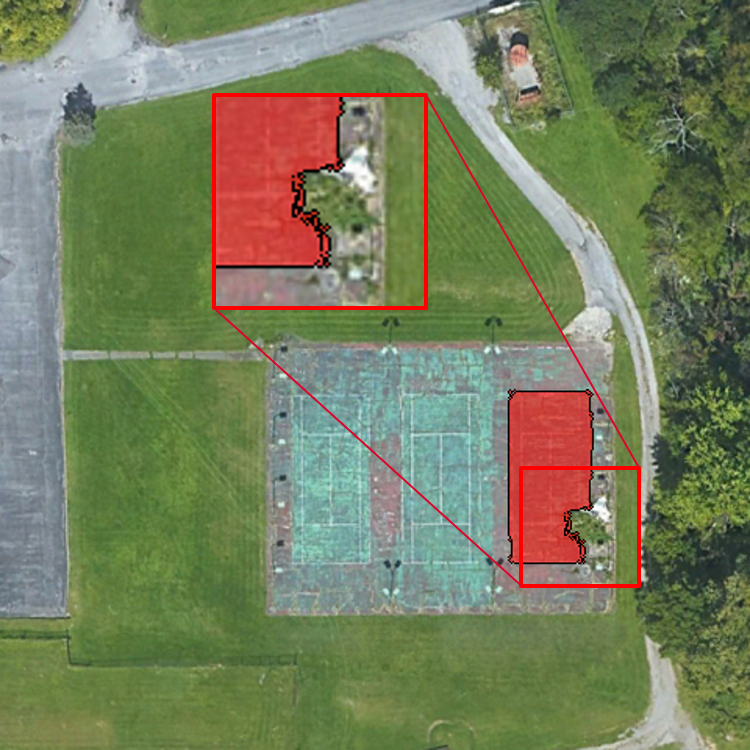}
  \end{subfigure}

  \vspace{-5mm}

  \begin{subfigure}[t]{1\linewidth}
  \subcaption*{\textbf{$Expression$:} The ship on the left}
  \end{subfigure}
  
  \begin{subfigure}[t]{0.19\linewidth}
  \includegraphics[width=0.95\linewidth]{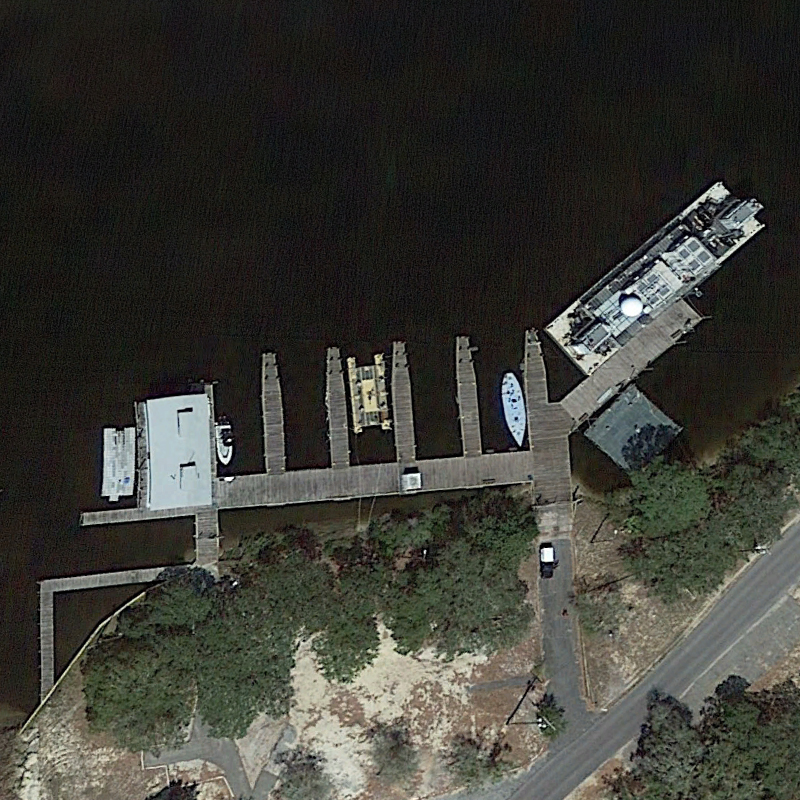}
  \subcaption*{Input Image}
  \end{subfigure}
  \begin{subfigure}[t]{0.19\linewidth}
  \includegraphics[width=0.95\linewidth]{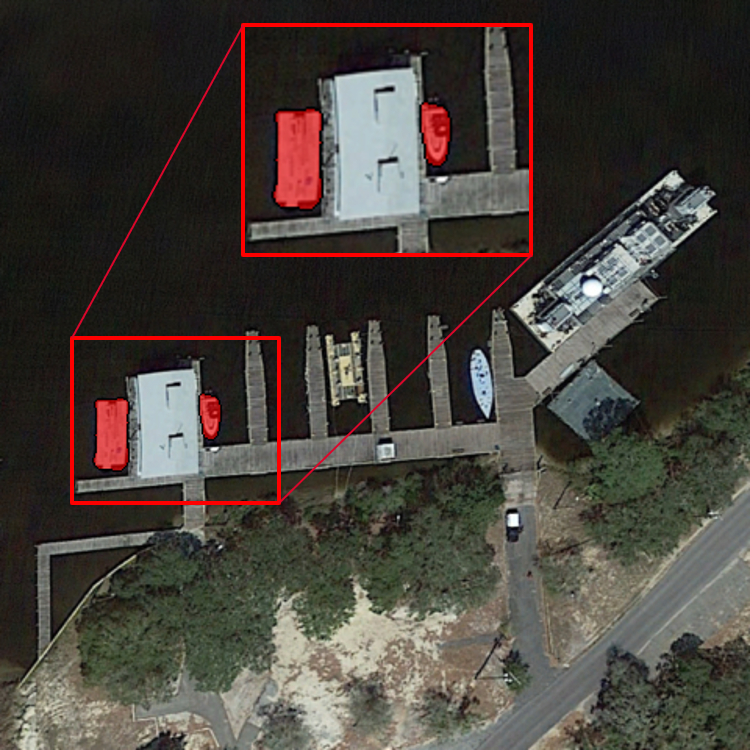}
  \subcaption*{w/o BAM}
  \end{subfigure}
  \begin{subfigure}[t]{0.19\linewidth}
  \includegraphics[width=0.95\linewidth]{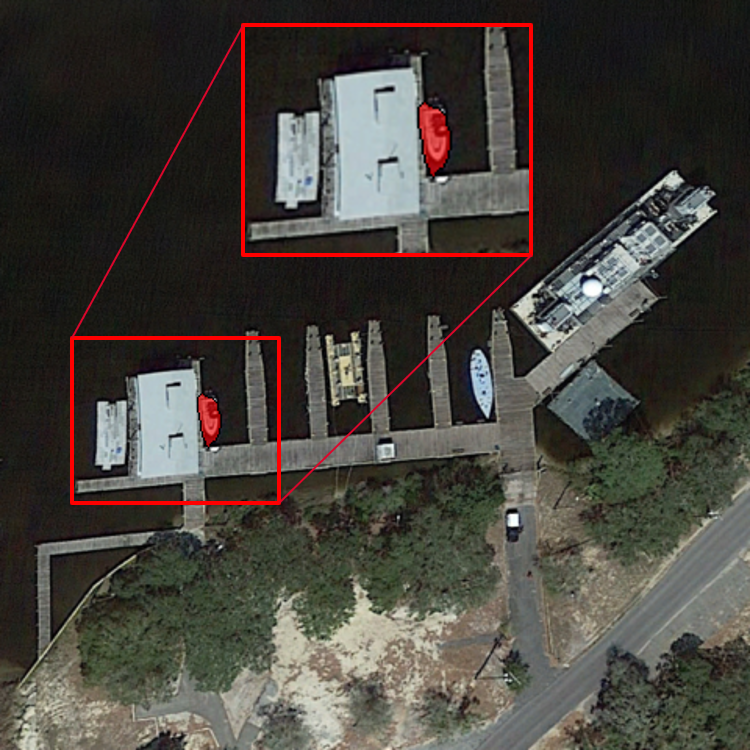}
  \subcaption*{w/o TCSA}
  \end{subfigure}
  \begin{subfigure}[t]{0.19\linewidth}
  \includegraphics[width=0.95\linewidth]{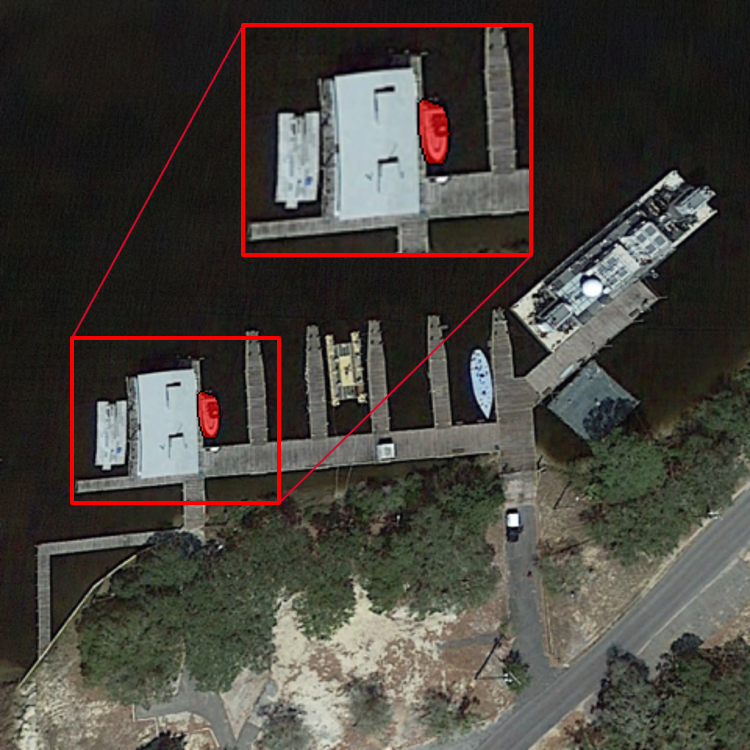}
  \subcaption*{SBANet}
  \end{subfigure}
  \begin{subfigure}[t]{0.19\linewidth}
  \includegraphics[width=0.95\linewidth]{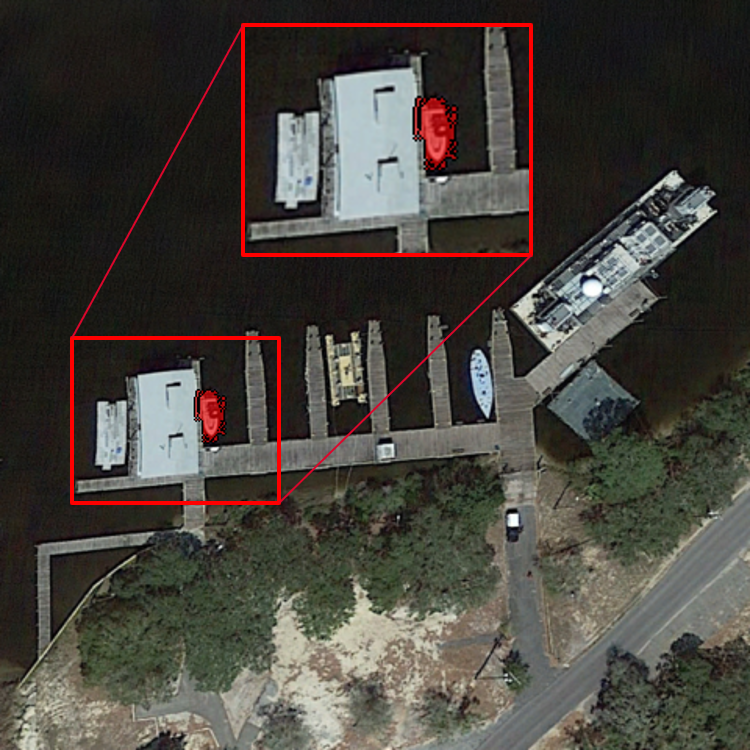}
  \subcaption*{Ground Truth}
  \end{subfigure}
  
  \vspace{-2mm}
  \caption{Visualized results for different components of the proposed SBANet on two examples from the RRSIS-D \cite{liu2024rmsin} dataset.
  }
  \vspace{-2mm}
  \label{fig: rrsis ablation}
\end{figure*}

\subsection{Qualitative Results}
\label{sec: qualitative}
To visually evaluate the effectiveness of our proposed SBANet, we provide qualitative comparison with previous methods on the two datasets.
In Fig.~\ref{fig: rrsis results}, we show several representative results on examples from the RRSIS-D \cite{liu2024rmsin} dataset.
With varying scales of referent objects in aerial images, our method achieved better segmentation performance with detailed pixel-wise predictions compared to the baseline LAVT \cite{yang2022lavt} and previous best-performing RMSIN \cite{liu2024rmsin} methods.
As shown from the first three rows, our method effectively and exactly captured linguistic focus based on the visual and textual context, thereby predicting precise masks for regional ground targets (especially when their boundaries are hard to be distinguished from the background).
For linear targets, our method also outperformed the others with completer and more consistent results as shown in the fourth row.
In more challenging cases, where multiple objects share the same category information, our method updates the linguistic features and suppresses the representation of irrelevant objects.
% For a more challenging case where multiple objects sharing the same category information are present, our method updated linguistic features and suppressed the representation of irrelevant objects.
For instance, from the last two rows, ``on the top" and ``blue" as cross-modal focuses play an important role in guiding the update of both linguistic and visual features.
We also present some examples from the RefSegRS \cite{yuan2024rrsis} dataset in Fig.~\ref{fig: refsegrs results}.
The results indicate that our method achieved more comprehensive and discriminative cross-scale information exchange and obtained more accurate masks compared to the baseline LAVT and previous best-performing LGCE \cite{yuan2024rrsis} methods.
Note that we used the same visualization method as the original benchmark works for the qualitative comparison.

Moreover, we show the ablation results on two core components of our SBANet in Fig.~\ref{fig: rrsis ablation}.
We selected two examples from the RRSIS-D dateset that include multiple objects of the same category.
For both examples, TCSA enabled the generation of complete masks with more precise boundaries.
Without BAM, we can see from the second row that the model failed to distinguish the target (\emph{e.g.,}~ship) from the platform on the most left.
These results further demonstrate the effectiveness of the modules proposed in SBANet for RRSIS.

\begin{figure}[t]
  \centering
  \begin{subfigure}[t]{1\linewidth}
  \subcaption*{\textbf{$Expression$:} The expressway service area is on the right of the vehicle on the left}
  \end{subfigure}

  \begin{subfigure}[t]{0.32\linewidth}
  \includegraphics[width=0.95\linewidth]{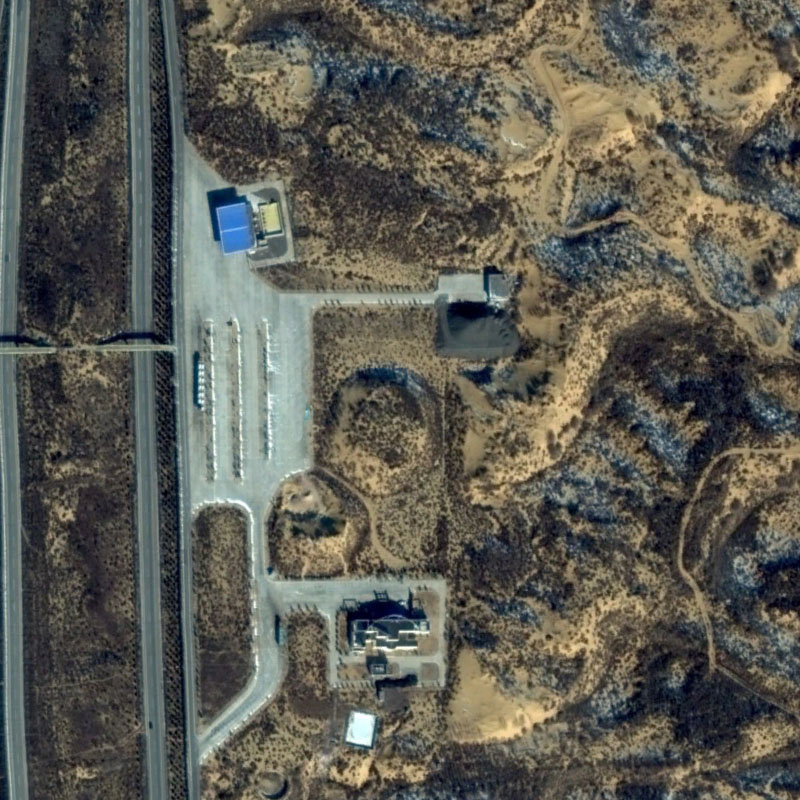}
  \end{subfigure}
  \begin{subfigure}[t]{0.32\linewidth}
  \includegraphics[width=0.95\linewidth]{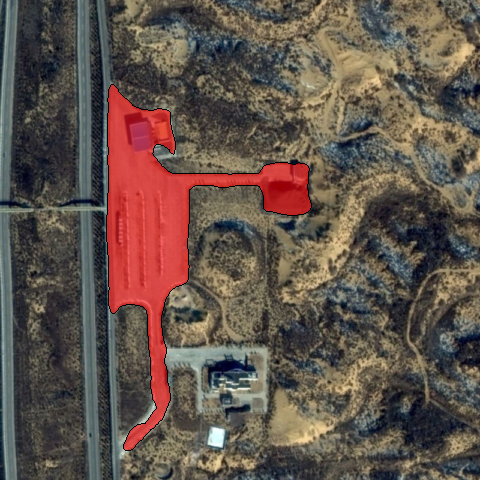}
  \end{subfigure}
  \begin{subfigure}[t]{0.32\linewidth}
  \includegraphics[width=0.95\linewidth]{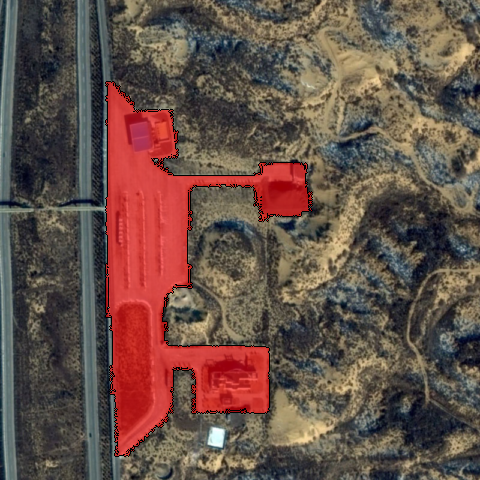}
  \end{subfigure}

  \vspace{-5mm}
  
  \begin{subfigure}[t]{1\linewidth}
  \subcaption*{\textbf{$Expression$:} The baseball field is on the lower right}
  \end{subfigure}

  \begin{subfigure}[t]{0.32\linewidth}
  \includegraphics[width=0.95\linewidth]{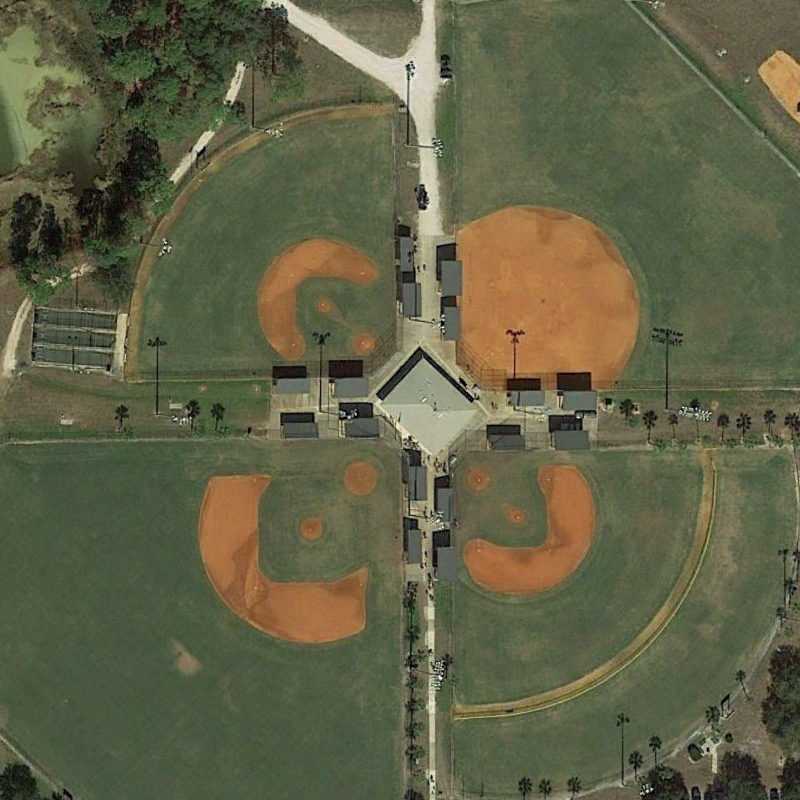}
  \subcaption*{Input Image}
  \end{subfigure}
  \begin{subfigure}[t]{0.32\linewidth}
  \includegraphics[width=0.95\linewidth]{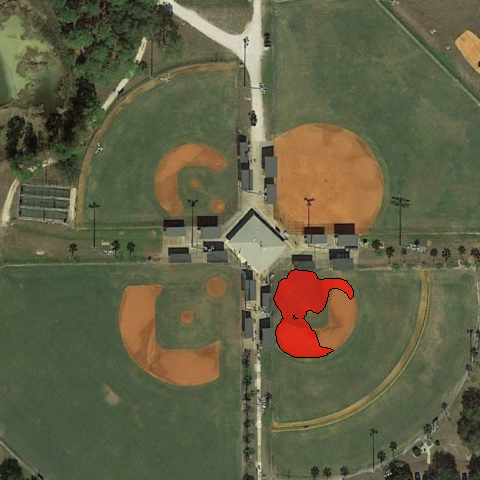}
  \subcaption*{SBANet (ours)}
  \end{subfigure}
  \begin{subfigure}[t]{0.32\linewidth}
  \includegraphics[width=0.95\linewidth]{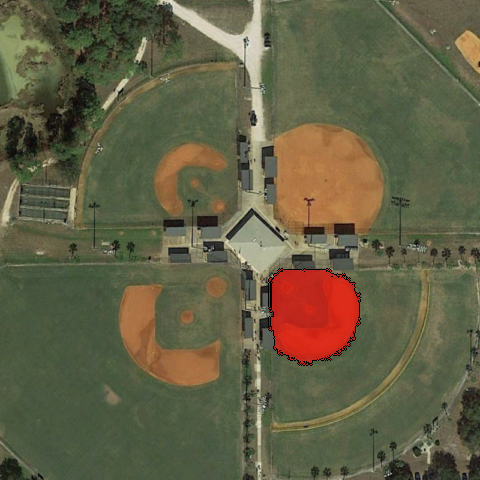}
  \subcaption*{Ground Truth}
  \end{subfigure}  
  
  \caption{Two failure cases from the RRSIS-D \cite{liu2024rmsin} dataset.}
  \label{fig: failure cases}
  \vspace{-3mm}
\end{figure}

For the failure cases shown in Fig.~\ref{fig: failure cases}, our approach failed to correctly extract the ``expressway service area" in the first example, primarily due to the low frequency of the concept and the ambiguity within the visual context.
These challenges hindered the model's ability to predict the building located at the bottom right.
The second example indicates that our model occasionally struggled to segment complete objects, particularly when they consist of multiple disconnected regions.
We believe that these failure patterns provide valuable insights and guidance for future research directions.

\section{Conclusion}
\label{sec: conclusion}
In this paper, we propose SBANet, a novel approach for referring remote sensing image segmentation.
Specifically, we introduce a bidirectional alignment module to update both visual and linguistic features for cross-modal interaction.
To selectively represent the visual context for the corresponding update, we design a dynamic feature selection block and learnable query tokens, respectively, for improving the alignment.
Furthermore, we propose a text-conditioned channel and spatial aggregator to enhance cross-scale information exchange with textually guided channel and spatial attentions.
Comprehensive experiments conducted on two benchmarks demonstrate the superiority of the proposed approach.

We expect that the new understanding of cross-modal alignment and the designed modules will benefit future research in this area.
Nevertheless, some issues can be pursued in the future to further promote the research.
First, the proposed approach can handle single referent object well based on unique query expressions following the standard referring image segmentation objective, while remote sensing scenarios often involve multiple objects sharing the same textual descriptions, which is more challenging.
Thus, we will improve SBANet's capability in capturing object-level information across entire input images to enhance the segmentation performance in the future work.
% Second, the refinement of features in the decoder should be further promoted, as its standard upsampling operations limit the model's performance in generating high-quality mask predictions.
Second, we aim to further explore the application of SBANet in multi-temporal and multi-source remote sensing research, including tasks such as change detection integrated with natural language expressions.

\bibliographystyle{IEEEtran}
\bibliography{refs.bib}

% \newpage

 

% \end{IEEEbiography}



\vfill

\end{document}